\documentclass[lettersize,journal]{IEEEtran}
\usepackage{tikz,xcolor}
\usepackage{amsmath,amsfonts}
\usepackage[caption=false,font=normalsize,labelfont=sf,textfont=sf]{subfig}
\usepackage{textcomp}
\usepackage{stfloats}
\usepackage{url}
\usepackage{verbatim}
\usepackage{cite}

\usepackage{color}
\usepackage{amsthm,bm,enumerate}
\usepackage[bottom]{footmisc}
\usepackage{setspace}
\usepackage{multirow}
\usepackage{graphicx}
\usepackage{epstopdf}
\usepackage{amsmath}
\usepackage{amsfonts}
\usepackage[colorlinks=true]{hyperref}
\hypersetup{urlcolor=blue, citecolor=blue,linkcolor=blue}
\usepackage{algorithm}
\usepackage{algorithmic}
\usepackage{array}
\usepackage{svg}

\definecolor{lime}{HTML}{A6CE39}
\DeclareRobustCommand{\orcidicon}{
\begin{tikzpicture}
\draw[lime, fill=lime] (0,0)
circle[radius=0.16]
node[white]{{\fontfamily{qag}\selectfont \tiny \.{I}D}};
\end{tikzpicture}
\hspace{-2mm}
}
\foreach \x in {A, ..., Z}{%
\expandafter\xdef\csname orcid\x\endcsname{\noexpand\href{https://orcid.org/\csname orcidauthor\x\endcsname}{\noexpand\orcidicon}}
}

\newtheorem{theorem}{Theorem}

\newcommand{\norm}[1]{\|{#1}\|}

\begin{document}
  \graphicspath{{./fig/}}
  
\title{A Robust Low-Rank Prior Model for Structured Cartoon-Texture Image Decomposition with Heavy-Tailed Noise}
\author{\IEEEauthorblockN{Weihao Tang\hspace{-1.5mm}\orcidA{} and Hongjin He\orcidB{}}
\thanks{
Corresponding author: Hongjin He (email: hehongjin@nbu.edu.cn).}
\thanks{The authors are with the School of  Mathematics and Statistics, Ningbo University, Ningbo 315211, China.}}

\markboth{Journal of \LaTeX\ Class Files,~Vol.~xx, No.~xx, March~2026}%
{W. Tang and H. He
}

\IEEEtitleabstractindextext{%
\begin{abstract}
Cartoon-texture image decomposition is a fundamental yet challenging problem in image processing. A significant hurdle in achieving accurate decomposition is the pervasive presence of noise in the observed images, which severely impedes robust results. To address the challenging problem of cartoon-texture decomposition in the presence of heavy-tailed noise, we in this paper propose a robust low-rank prior model. Our approach departs from conventional models by adopting the Huber loss function as the data-fidelity term, rather than the traditional $\ell_2$-norm, while retaining the total variation norm and nuclear norm to characterize the cartoon and texture components, respectively. Given the inherent structure, we employ two implementable operator splitting algorithms, tailored to different degradation operators. Extensive numerical experiments, particularly on image restoration tasks under high-intensity heavy-tailed noise, efficiently demonstrate the superior performance of our model. 
\end{abstract}

}

\maketitle

\IEEEdisplaynontitleabstractindextext

\IEEEpeerreviewmaketitle

\section{Introduction}\label{sec_intro}

\IEEEPARstart{C}{artoon-texture} image decomposition, which aims to split an observed image into a piece-wise smooth cartoon component that captures large-scale geometric structures, and a highly-oscillatory texture component that encodes fine-scale repetitive patterns, is a fundamental task in modern image processing. This separation enables dedicated processing of each part and has proven crucial for applications such as denoising, inpainting, super-resolution, compression, and material analysis (e.g., see \cite{BVSO03,FSBM10,Meyer01}).
Mathematically, given an observed image $b_{0}\in\mathbb{R}^{m\times n}$, one seeks
\begin{equation}\label{eq:model}
b_{0} = \Phi(u+v) + \varepsilon,
\end{equation}
where $u$ denotes the cartoon part, $v$ represents the texture component, $\Phi$ is a known linear operator (which can be the identity operator or other degradation operators, such as blurring and down-sampling operations), and $\varepsilon$ is an unknown noise term.  

Generally speaking, solving the inverse problem \eqref{eq:model} presents significant challenges due to its underdetermined nature and severe ill-posedness, even in noise-free and pure image decomposition (i.e., $\Phi=I$, where $I$ represents the identity operator) scenarios. For practical applications, the key to successful decomposition from degraded measurements $b_0$ lies in the judicious regularization of $u$ and $v$, incorporating suitable prior knowledge about their properties. As shown in \cite{FSBM10}, sparsity and compressibility assumptions often yield effective regularizations. For this reason, we restrict our attention to structured images in which the cartoon and texture components can be accurately characterized by widely used regularizers.

In the literature, the seminal regularizer for cartoon part is the well-known Total-Variation (TV) semi-norm (see \cite{ROF92}), which is popular for its ability to preserve discontinuities (sharp edges) in images. However, direct application of the TV model \cite{ROF92} lacks an explicit texture prior and tends to leave low-frequency textures in the residual $v$. Consequently, Meyer~\cite{Meyer01} introduced the so-named dual $G$-norm to promote the oscillatory texture, which successfully extracts textures, yet the non-smoothness of $G$-norm makes the underlying optimization problem hard to solve. Subsequently, Vese and Osher~\cite{VO03} proposed the use of approximate dual norms and wavelet/frame-based priors, though this approach comes at the expense of either theoretical elegance or computational efficiency. Then, Ng et al. \cite{NYZ13} introduced a structured optimization and an efficient multi-block splitting method to improve the quality of the decomposition.

In reality, many natural images, such as facades, fabrics, or wallpapers, exhibit regular patterns that can be modeled as approximately low-rank matrices. Then, Schaeffer and Osher~\cite{SO13} judiciously translated this observation into the Low-Patch-Rank (LPR) prior and established the following decomposition model for \eqref{eq:model}:
\begin{equation}\label{LRP}
\min_{u,v}\;\tau\| |\nabla u| \|_{1}+\mu\|\mathcal{P}v\|_{*}+\frac{1}{2}\left\|\Phi(u+v)-b_{0}\right\|^{2},
\end{equation}
where $\nabla$ denotes the first-order derivative operator,  $\mathcal{P}$ extracts overlapping patches and $\|\cdot\|_{*}$ denotes the nuclear norm. In many cases, the LPR model \eqref{LRP} is effective when the texture is coherent across the whole image, but becomes sub-optimal for images with multiple, locally different patterns.
Therefore, Ono et al. \cite{OMY14} proposed the Block Nuclear-Norm (BNN) prior that imposes low-rank constraints on local blocks instead of global patches.  
While BNN improves decomposition quality, its block-wise structure increases algorithmic complexity and, more importantly, prevents a straightforward extension to color images.
Recently, Zhang and He \cite{ZH21} introduced the Customized Low-Rank Prior (CLRP) model for some globally well-patterned images:
\begin{equation}\label{CLRP}
\min_{u,v}\;\tau\| |\nabla u| \|_{1}+\mu\| v\|_{*}+\frac12\left\| \Phi(u+v)-b_{0}\right\|^{2},
\end{equation}
which directly regularizes the entire texture component by the global nuclear norm and couples it with the classical TV cartoon prior. As demonstrated in \cite{ZH21}, the CLRP model \eqref{CLRP} is strikingly simple, parameter-efficient, and naturally extends to multi-channel images. The most recent work \cite{LWC25} leverages iterative weighted least square and low-rank regularization to separately model cartoon and texture components. This approach is able to effectively enhance edges and suppress texture through iterative updates of an edge-preserving weight matrix.
In recent years, deep learning-based image decomposition models have become a research hotspot. With the nonlinear fitting ability of neural networks, these models learn decomposition mapping via end-to-end training, breaking the limitation of manual regularization in traditional models. A typical work is the PnP-based decomposition method \cite{GAT25}, which replaces regularization with deep denoisers to balance flexibility and performance. CNN, GAN and other networks are also directly used for Cartoon-Texture decomposition to achieve end-to-end output.

\begin{figure*}[!ht]
	\centering
	\includegraphics[width=0.9\textwidth]{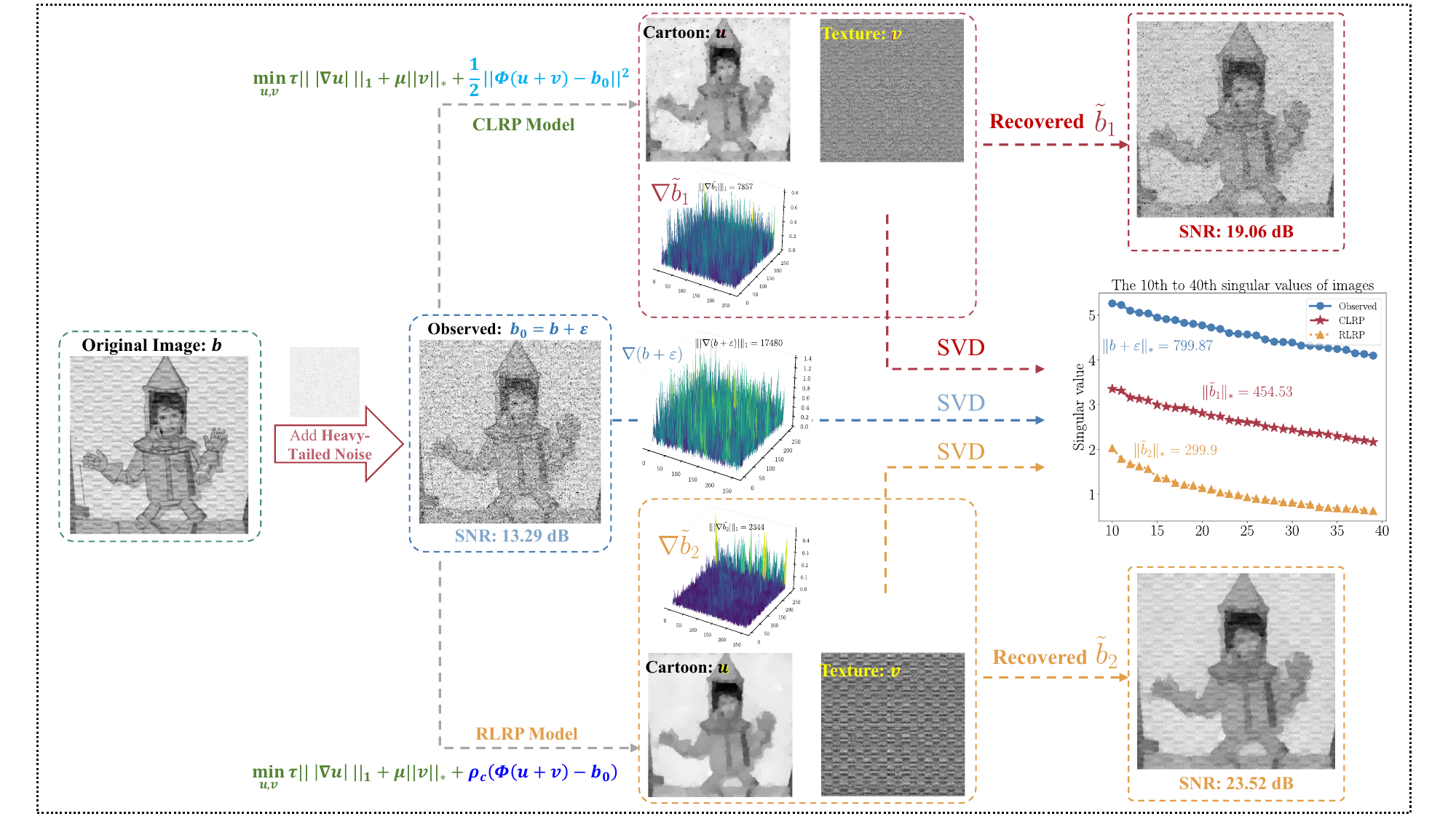}
	\caption{A comparison between the CLRP model \eqref{CLRP} and our RLRP model \eqref{RLRP}, which provides a conceptual explanation of the robustness of our RLRP model to heavy-tailed noise.}
	\label{motivation}
\end{figure*}

Note that the aforementioned approaches typically perform well in scenarios involving no noise or white Gaussian noise. Consequently, the quadratic data fidelity term derived from the $\ell_2$ norm is commonly employed to handle such noise. However, in practice, images are often corrupted by heavy-tailed or impulsive noise (\cite{MMCAB19,JC15,Yan12}), rendering the traditional quadratic data fidelity term highly sensitive and unstable.
As illustrated in Fig. \ref{motivation}, heavy-tailed noise significantly disrupts both gradient sparsity and low-rank properties, which encourages us to search more robust alternatives.

Inspired by the success of the Huber function in robust regression for data fitting (widely used in machine learning and regression analysis, see \cite{Huber73,SZF20}), we adopt it to formulate the data fidelity term in this paper. Promisingly, the Huber function is a piecewise function that adaptively balances the sensitivity of the $\ell_2$-norm for moderate noise and the robustness of the $\ell_1$-norm for outliers, providing enhanced resilience against diverse noise distributions. Building on this, we propose a novel model for decomposing an observed image corrupted by heavy-tailed noise into its cartoon and texture components. Our key contributions are threefold:
\begin{itemize}
	\item We introduce a Robust Low-Rank Prior (RLRP) model that explicitly addresses heavy-tailed noise via the Huber loss. It is noteworthy that the Huber function incorporates an adaptive threshold, enabling a dynamic selection between the $\ell_1$-norm and $\ell_2$-norm. For pixels with noise close to zero (near the mean), it behaves like the $\ell_2$-norm, preserving sensitivity to small deviations. For pixels heavily corrupted by outliers, where values deviate significantly from local regions, it switches to the $\ell_1$-norm, aligning with low-rank and smoothness priors. Thus, rather than enforcing strict data fidelity, our model prioritizes regularization-driven estimation.
	\item We develop globally convergent and efficient optimization algorithms for both identity $\Phi = I$ and general linear degradation operators $\Phi \neq I$. Due to the piece-wise structure of the Huber function, the appearance of general operator $\Phi$ makes our RLRP model more difficult than traditional models. Therefore, we introduce two customized algorithms so that they are easily implementable with closed-form solutions. Specifically, we employ the partially parallel splitting algorithm \cite{HHY16} with closed-form proximal updates to deal with the case $\Phi=I$. For general operators $\Phi\neq I$, we derive a primal-dual hybrid gradient algorithm \cite{CP11} with parallel and explicit updates, overcoming the challenges posed by the piecewise Huber loss in numerical optimization.  
	\item Through extensive experiments on synthetic and real-world images, we demonstrate that our RLRP model significantly outperforms traditional Gaussian-noise-based models when images are corrupted by heavy-tailed noise.  
\end{itemize}

This paper is organized as follows. In Section \ref{sec_pre}, we first introduce the notations that will be used throughout the paper. In Section \ref{sec:model-alg}, we propose a novel robust low-rank prior model for \eqref{eq:model}. Then, we present two implementable algorithms for our model. In Section \ref{sec_exp}, we conduct the numerical performance of our approach (i.e., model and algorithms) on handling the heavy-tailed noise. Finally, we complete this paper with some concluding remarks in Section \ref{sec_con}.

\section{Notations}\label{sec_pre}
In this section, we recall some notations that will be used in this paper.

 Let $\mathbb{R}^n$ represent an $n$-dimensional Euclidean space. 
For a given $x\in {\mathbb R}^n$, we denote $\|x\|_{p}$ as the $\ell_p$ norm, i.e., $\|x\|_p:=(\sum_{i=1}^{n}|x_i|^p)^{\frac{1}{p}}$ for $1\leq p <\infty$, 
where $x_i$ is the $i$th component of vector $x$. In particular, we will use $\|\cdot\|$ to denote the standard $\ell_2$-norm for notational simplicity.  
When $p=\infty$, it represents the infinity-norm defined by $ \|x\|_\infty:=\max_{i} |x_i| $.  Given a matrix $X\in \mathbb{R}^{m\times n}$,  we denote its nuclear norm by $\|X\|_*:= \sum_{i=1}^r \sigma_i(X)$, 
where $\sigma_i(X)$ is the $i$th largest singular value of $X$ and $r := \min(m,n)$. 
In addition, the Frobenius norm of $X$ is defined by $\|X\|_F:=\sqrt{\sum_{i=1}^m\sum_{j=1}^{n}X_{ij}^2}$. 
Note that the Frobenius norm can be regarded as the $\ell_2$-norm of vector that is stacked by a matrix,  
so the $\|\cdot\|$ throughout represents the $\ell_2$-norm of a vector or the Frobenius norm of a matrix for notational simplicity. 
The first-order derivative operator is denoted as $\nabla:=\left(\begin{array}{c} \nabla_x \\ \nabla_y
\end{array}\right).$  
Then, in our experiments, we will use the isotropic TV defined by $(|\nabla u|)_i:=\sqrt{(\nabla_x u)_i^2+(\nabla_y u)_i^2}$ (see \cite{ROF92}). 
Throughout this paper, $I$ and $\textbf{0}$ represent the identity matrix (or operator) and zero vector for simplicity.

Below, we recall the proximity operator that will be used in our algorithmic implementations. For a given function $f:\mathbb{R}^n\to\mathbb{R}\cup \{+\infty\}$, $\text{dom}f$ represents its domain, and its conjugate function is defined by $  f^*(y):=\sup_{x\in\text{dom}f}(y^\top x-f(x)) $. Then, we define
\begin{equation}\label{def:prox}
\text{prox}_{\beta f}(\bm{a}):=\arg\min_{x\in\text{dom}f} \left\{f(x)+\frac{1}{2\beta}\|x-\bm{a}\|^2\right\}
\end{equation}
as the proximity operator of $f$, where $\beta>0$. Consequently, when $f$ is specified as the indicator function associated with the set consisting of elements with bounded infinity-norm $\mathbb{B}^{\infty}_\tau=\left\{x\;:\;\max_{i}|x_i|<\tau\right\}$ with a given $\tau>0$, then the proximity operator \eqref{def:prox} corresponds to the projection operator as follows: 
\begin{equation}\label{clip_op}
\mathrm{Clip}_{\tau}(\bm{a}):=\mathrm{sign}(\bm{a})\cdot\min(|\bm{a}|,\tau),  \;\; \text{for}\;\;\bm{a}\in\mathbb{B}^{\infty}_\tau,
\end{equation}
where `sign($\cdot$)' is the sign function. Moreover, when $f$ is taken as $f(x)=\|x\|_1$, the proximity operator \eqref{def:prox} reduces to the well-known shrinkage operator ``shrink($\cdot$,$\cdot$)" defined by
\begin{equation}\label{shrinkv}
\mathrm{shrink}(\bm{a},\tau) := {\rm sign}(\bm{a}) \cdot \max\{|\bm{a}|-\tau,\;0\},
\end{equation}
The shrinkage operator is also available for matrices. 
Given a matrix $X$ and a scalar $\tau>0$, let $X = U \Sigma V^\top$ be the singular value decomposition (SVD) of $X$, 
the soft-thresholding operator (see \cite{cai2010singular}) ${\mathcal S}: \mathbb{R}^{m\times n} \to \mathbb{R}^{m \times n}$ is defined as
\begin{equation}\label{shrinkm}
{\mathcal{S}}(X,\tau) = U \mathrm{shrink}(\Sigma,\tau) V^\top.
\end{equation}
To end this section, we recall the Huber function $\rho_{c}(x)$, which is defined as:
\begin{equation}\label{huber}
\rho_c(x)=\left\{\begin{array}{ll}
\frac{1}{2}x^2, & \text{if $|x|\leq c$},\\
c|x| - \frac{c^2}{2}, & \text{if $|x|> c$},
\end{array}\right.
\end{equation}
where $x\in\mathbb{R}$ and $c>0$ is a hyperparameter that determines the threshold for switching between squared and absolute loss. Interestingly,  we can apply the Huber function element-wise to the entries of the matrix $X\in\mathbb{R}^{m\times n}$ and then aggregates the results, i.e.,
$ \rho_c(X) = \sum_{i=1}^{m}\sum_{j=1}^{n}\rho_c(X_{ij}) $.
Note that such a function still enjoys the smoothness at $x=0$, which will be helpful in optimization. Besides, the conjugate function of the Huber function is defined as:
\begin{equation}\label{huber_con}
\rho_c^*(x)=\left\{
\begin{array}{ll}
\frac{1}{2}x^2, & \text{if $|x|\leq c$}, \\
+\infty, & \text{if $|x|>c$}.
\end{array}
\right.
\end{equation}

\section{The RLRP Model and Algorithms}\label{sec:model-alg}
In this section, we first introduce a novel Robust Low-Rank Prior (RLRP) model for image decomposition. Then, we reformulate our proposed RLRP model as some structured optimization problems so that we can easily employ efficient optimization algorithms to obtain ideal solutions of \eqref{eq:model}. 

\subsection{The RLRP model}

As demonstrated in \cite{ZH21}, texture components can be effectively characterized by a low-rank function when images contain large and well-preserved texture regions. A widely used convex surrogate for low-rank functions is the nuclear norm, which has proven highly successful in image and video processing. Meanwhile, the TV norm is an ideal choice for regularizing the cartoon component due to its ability to capture the smooth continuity of the background. Following the approach of \cite{ZH21}, we therefore adopt the TV norm and nuclear norm to regularize the cartoon and texture parts, respectively.

In the literature, the standard $\ell_2$-norm has been shown to effectively handle Gaussian noise. 
However, for heavy-tailed noise (e.g., t-distributed noise), 
the quadratic growth of the $\ell_2$-norm often leads to suboptimal performance.  Consequently, the traditional $\ell_2$-norm is not ideal for heavy-tailed noise (see Fig. \ref{motivation}), 
as further evidenced by our numerical experiments (Section \ref{sec_exp}). To address this, we replace the $\ell_2$-norm with the Huber function for the data fidelity term. Specifically, our Robust Low-Rank Prior (RLRP) model is formulated as:
\begin{equation}\label{RLRP}
\min_{u,v} \quad \tau \||\nabla u|\|_1 + \mu \|v\|_* + \rho_c(\Phi(u+v)-b_0),
\end{equation}
where $\rho_c(\cdot)$ is defined by \eqref{huber}, $\tau$ and $\mu$ are positive regularization parameters balancing the contributions of the cartoon and texture components.

It is easy to see from \eqref{RLRP} that the piece-wise structure of the Huber function and the coupling of $u$ and $v$ through the degraded operator $\Phi$ make directly solving \eqref{RLRP} nontrivial. Besides, when $\Phi=I$ or $B$ being the blurring operator, the related subproblems usually can be solved with low computational cost after a {\it Fast Fourier Transformation} (FFT). However, when $\Phi=S$ being the down-sampling operator, which cannot be transferred into a diagonal form, thereby making the related subproblems more difficult. Thus, effectively exploiting the structure of \eqref{RLRP} is essential for devising an efficient algorithm. To address this, we below develop two tailored algorithms for the cases $\Phi=I$ and $\Phi\neq I$, respectively.

\subsection{Solving model \eqref{RLRP} with $\Phi=I$}\label{sec:IPhi}

Note that the objective function of \eqref{RLRP} is the sum of three convex functions, and each of them has their own structure. Moreover, the last Huber loss function has coupled variables $u$ and $v$. Therefore, we introduce two auxiliary variables $y$ and $z$ to fully separate the three objectives. Specifically, we let $\nabla u=y$ and $\Phi(u+v)=z$ with $\Phi=I$, then \eqref{RLRP} can be recast as
\begin{equation}\label{R-RLRP}
\begin{aligned}
\min\quad &\mu \Vert v \Vert_*+\tau \Vert |y| \Vert_1 + \rho_{c}(z-b_0) \\
\text{s.t.}\quad &\left(\begin{matrix}
\nabla \\
I
\end{matrix}
\right) u + 
\left(\begin{matrix}
O \\
I
\end{matrix}
\right) v + 
\left(\begin{matrix}
-I  & O\\
O & -I
\end{matrix}
\right) \left(\begin{matrix}
y \\
z
\end{matrix}
\right) =0,
\end{aligned}
\end{equation}
where $O$ represents the zero matrix. Consequently, by letting
$$A =\left(\begin{matrix}
\nabla \\
I
\end{matrix}
\right),\; B =\left(\begin{matrix}
O \\
I
\end{matrix}
\right) ,\;  C =\left(\begin{matrix}
-I  & O\\
O & -I
\end{matrix}
\right),\; w =\left(\begin{matrix}y \\z\end{matrix}\right),$$
the model \eqref{R-RLRP} falls into a standard linearly constrained three-block separable convex minimization problem:
\begin{equation}\label{tb}
\begin{split}
\min_{u,v,w} \quad & f(u) + g(v) + h(w)\\
\mathrm{s.t.}\quad & Au+Bv+Cw = 0,
\end{split}
\end{equation}
where
\begin{equation*}
\begin{cases}
f(u)=0,\\ 
g(v)=\mu\|v\|_*,\\ 
h(w)=\tau \||y|\|_1 +\rho_{c}(z-b_0).
\end{cases}
\end{equation*}
Now, we first construct the augmented Lagrangian function:
\begin{align}\label{alag}
\mathcal{L}_\beta(u,v,w,\lambda) =\;& f(u) + g(v)+h(w) \nonumber\\ &-\lambda^\top(Au+Bv+Cw) \nonumber\\
&+\frac{\beta}{2}\left\| Au+Bv+Cw\right\|^2,
\end{align}
where $\beta>0$ is a penalty parameter, 
and $\lambda$ is called Lagrangian multiplier associated to the linear constraints in \eqref{tb}. Then, by the algorithmic framework introduced in \cite{HHY16}, we follow the order $u\to \lambda \to v\to w$ to make a prediction step. Specifically, we show their explicit iterative schemes as follows.
\begin{itemize}
	\item Update $u^{k+1}$ via 
	\begin{align*}
	u^{k+1}=&\arg\min_u \left\{f(u)+\frac{\beta}{2s}\Vert Au+{\bm a}_u^k\Vert^2\right\}\\
	=&\arg \min_u  \frac{\beta}{2s}\left\{\left\| \nabla u - y^k - \frac{s}{\beta}\lambda_1^k\right\|^2 \right.\\
	&\qquad\qquad\;\;\; \left.+ \left\| u + v^k - z^k-\frac{s}{\beta}\lambda_2^k \right\|^2\right\}.
	\end{align*}
	where $\bm{a}_u^k =Bv^k+Cw^k-\frac{s}{\beta}\lambda^k $. As a consequence, by the first-order optimality condition, we easily obtain
	\begin{align}\label{u_update}
	(\nabla^\top\nabla+I)u^{k+1}=&\nabla^\top\left(y^k+\frac{s}{\beta}\lambda_1^k\right) \nonumber \\
	&+\left(z^k- v^k+\frac{s}{\beta}\lambda_2^k\right).
	\end{align}	
	\item Compute a prediction $\tilde{\lambda}^k$ via 
	\begin{equation}\label{ppsml}
	\tilde{\lambda}^k=\lambda^k-\frac{\beta}{s}\left(Au^{k+1}+Bv^k+Cw^k\right).
	\end{equation}
	By the block structure of $A,B,C$, we let $\lambda=(\lambda_1^\top,\lambda_2^\top)^\top$, then   \eqref{ppsml} can be rewritten as
	\begin{equation}\label{prel}
\left(\begin{matrix}
	\tilde{\lambda}_1^k \\
	\tilde{\lambda}_2^k\\
	\end{matrix}
	\right)=\left(\begin{matrix}
	\lambda_1^k - \frac{\beta}{s}\left(\nabla u^{k+1}-y^k\right) \\
	\lambda_1^k - \frac{\beta}{s}\left(u^{k+1}+v^{k+1}-z^k\right)
	\end{matrix}
	\right).
	\end{equation}
	\item Generate $\tilde{v}^k$ and $\tilde{w}^k$ simultaneously via 
	\begin{equation}\label{ppsmvw}
	\begin{cases}
	\tilde{v}^k=\arg\min_v \left\{g(v)+\frac{r \beta}{2}\norm{Bv-\bm{a}_v^k}^2\right\}, \\
	\tilde{w}^k=\arg\min_w \left\{h(w)+\frac{r \beta}{2}\norm{Cw-\bm{a}_w^k}^2\right\},
	\end{cases}
	\end{equation}
	where 
	\begin{equation*}
	\begin{cases}
	\bm{a}_v^k = v^k+\frac{1}{r\beta}(2\tilde{\lambda}_2^k-\lambda_2^k), \\ \bm{a}_w^k =\left(\begin{array}{c}
	\bm{a}_y^k \\ \bm{a}_z^k
	\end{array}\right)= \left(\begin{array}{c}
	- y^k + \frac{1}{r\beta}\left(2\tilde{\lambda}_1^k - \lambda_1^k\right) \\
	-z^k+\frac{1}{r\beta}\left(2\tilde{\lambda}_2^k-\lambda_2^k\right)
	\end{array}\right). 
	\end{cases}
	\end{equation*}
	By invoking the fact $g(v)=\mu \|v\|_*$, the prediction on $\tilde{v}^k$ in \eqref{ppsmvw} is specified as
	\begin{align}\label{v_update}
	\tilde{v}^k &=  \arg \min_v \left\{\mu \norm{v}_* + \frac{r\beta}{2}\norm{v-\bm{a}_v^k}^2\right\} \nonumber \\
	&=\mathcal{S}\left(\bm{a}_{v}^k,\frac{\mu}{r\beta}\right).
	\end{align}
	where $\mathcal{S}(\cdot,\cdot)$ is given in \eqref{shrinkm}. With the notation $h(w)$, the prediction on $\tilde{w}^k$ in \eqref{ppsmvw} reads as
	$$
	\begin{aligned}
	\tilde{w}^{k}=\arg \min_w \tau \Vert |y| \Vert_1 + \rho_{c}(z-b_0)
	+\frac{r\beta}{2}
	\left\|
	\left(
	\begin{matrix}
	y+\bm{a}_{y}^k\\
	z +\bm{a}_{z}^k
	\end{matrix}
	\right)
	\right\|^2.
	\end{aligned}
	$$
	By the separability of the $h(w)$ with respect to $y$ and $z$, the above subproblem is separated into two subproblems on $y$ and $z$. So, we can also update $\tilde{y}^k$ and $\tilde{z}^k$ in a parallel way.
	\begin{itemize}
		\item First, the prediction $\tilde{y}^k$ reads as
		\begin{align}\label{y_update}
		\tilde{y}^k = & \arg \min_y \left\{\tau \norm{|y|}_1 +\frac{r\beta}{2}\norm{y+\bm{a}_y^k}^2\right\} \nonumber \\
		= & \mathrm{shrink}\left(-\bm{a}_y^k\;,\;\frac{\tau}{r\beta}\right).
		\end{align}
		where $\mathrm{shrink}(\cdot,\cdot)$ is defined by \eqref{shrinkv}.
		\item Similarly, the prediction $\tilde{z}^k$ arrives at
		\begin{align}\label{z_update}
		\tilde{z}^k &=  \arg \min_z \left\{\rho_c(z-b_0)+\frac{r\beta}{2}\norm{z+\bm{a}_z^k}^2\right\}\nonumber \\
		&=\mathrm{prox}_{\frac{1}{r\beta}\rho_{c}(\cdot)}(-b_0-\bm{a}_z^k)+b_0,
		\end{align}
		where the proximity operator of Huber function is defined as
		$$
		\text{prox}_{\beta\rho_c(\cdot)}(x)=\left\{\begin{array}{ll}
		\text{prox}_{\frac{\beta}{2}\|\cdot \|}(x), & \text{if $|x|\leq c$}\\
		\text{prox}_{c\beta\|\cdot\|_1}(x),& \text{if $|x|>c$}.
		\end{array}\right.
		$$
	\end{itemize}
	\item Finally, we make a correction on $(v^{k+1},w^{k+1},\lambda^{k+1})$ via a relaxation step:
	\begin{equation}\label{relax}
	\bm{\zeta}^{k+1}=\left(\begin{array}{c} v^{k+1} \\ w^{k+1} \\ \lambda^{k+1}
	\end{array}\right) =\left(\begin{array}{c} v^{k} \\ w^{k} \\ \lambda^{k}
	\end{array}\right) - \gamma \left(\begin{array}{c} v^{k}- \tilde{v}^k \\ w^{k} -\tilde{w}^k \\ \lambda^{k}-\tilde{\lambda}^k
	\end{array}\right).
	\end{equation}
\end{itemize}
With the above derivations of all subproblems, we can summarize them into Algorithm \ref{alg_I}.

\renewcommand{\algorithmicrequire}{\textbf{Require:}}
\begin{algorithm}[!htbp]
	\setstretch{1.1} 
	\caption{Solving~problem~\eqref{RLRP} when $\Phi=I$.} \label{alg_I}
	\begin{algorithmic}[1]
		\REQUIRE Set initial points $(v^0,y^0,z^0,\lambda_1^0,\lambda_2^0)$, penalty parameter $\beta>0$, $\gamma \in (0,2)$, $r$ and $s$ satisfying $rs>2$, and a  tolerance $\epsilon >0$.
		\FOR{$k=0,1,\cdots$}
		\STATE Update $u^{k+1}$ via \eqref{u_update}.
		\STATE Compute $\tilde{\lambda}_1^k$ and $\tilde{\lambda}_2^k$ via \eqref{prel}.
		\STATE Predict $\tilde{v}^k$ via \eqref{v_update}.
		\STATE Generate $\tilde{y}^k$ and $\tilde{z}^k$ via \eqref{y_update} and \eqref{z_update}, respectively.
		\STATE Correct $(v^{k+1},w^{k+1},\lambda^{k+1})$ via  \eqref{relax}.
		\ENDFOR
	\end{algorithmic}
\end{algorithm}

Note that \eqref{RLRP} is equivalent to the following saddle point problem:
\begin{equation}\label{SD-one}
\max_{\lambda}\min_{u,v,w}\mathcal{L}_0(u,v,w,\lambda),
\end{equation}
where $\mathcal{L}_0$ is given by \eqref{alag} with setting $\beta=0$. Consequently, by letting 
$$ Q =\left(\begin{matrix}
r\beta B^\top B & O & -B^\top\\
O & r\beta C^\top C & -C^\top\\
-B & -C & \frac{s}{\beta}I
\end{matrix}\right),  $$
it follows from \cite[Theorem 3.2]{HHY16} that Algorithm \ref{alg_I} is globally convergent and has the worst-case $\mathcal{O}(1/N)$ convergence rate under the condition $rs>2$ (which ensures the positive definiteness of $Q$). Here, we skip its proof for the conciseness of this paper. The readers who are concerned with its details can refer to \cite{HHY16}.

\begin{theorem}\label{theorem_1}
Let $\Omega^*$ be the solution set of \eqref{SD-one} and $\Delta^*:=\left\{\zeta^*=(v^*,w^*,\lambda^*)\;|\;\xi^*\in\Omega^*\right\}$, where $\xi^*=(u^*, v^*, w^*, \lambda^*)$. Then, the sequences $\{\xi^k\}$, $\{\zeta^k\}$, and $\{\bar{\zeta}^k\}$ generated by Algorithm \ref{alg_I} satisfy the following assertions:
\begin{enumerate}
    \item[(a).]  The sequence of subvectors $\{\zeta^k\}$ of $\{\xi^k\}$ converges globally to a member of $\Delta^*$.
    \item[(b).]  For any integer number $N>0$, we have 
    \begin{equation*}
           \left\| \zeta^N - \tilde{\zeta}^N \right\|_Q^2 \le \frac{1}{\gamma(2-\gamma)(N+1)}\left\| \zeta^0-\zeta^*\right\|_Q^2,
    \end{equation*}
    where the Q-norm of $x\in\mathbb{R}^n$ is defined by 
    $\norm{x}_Q:=\sqrt{x^{\top}Qx}$.
\end{enumerate}
\end{theorem}

\subsection{Solving model \eqref{RLRP} with $\Phi\neq I$}\label{sec:GPhi}
As discussed in Section \ref{sec:IPhi}, when $\Phi$ differs from the identity operator, the $u$-subproblem no longer retains the simplicity of \eqref{u_update}. Moreover, the $v$-subproblem will lose its explicit form of the proximity operator. In this case, the aforementioned partially parallel splitting method \cite{HHY16} is not necessarily the best choice for $\Phi\neq I$. Therefore, we first reformulate \eqref{RLRP} as a structured convex-concave saddle point problem, which is beneficial for employing the popular first-order primal-dual algorithm \cite{CP11} to solve the problem under consideration. 

First, by letting $\nabla u=x$ and $\Phi(u+v)=z$, we can rewrite \eqref{RLRP} as 
\begin{equation}\label{dual_pro}
\begin{aligned}
\min_{u,v,x,z} \quad&\tau \Vert |x| \Vert_1 + \mu \Vert v \Vert_* + \rho_{c}(z-b_0) \\
\text{s.t.}\quad& \left(\begin{matrix}
\nabla \\
\Phi
\end{matrix}
\right) u + 
\left(\begin{matrix}
O \\
\Phi
\end{matrix}
\right) v - 
\left(\begin{matrix}
x  \\
z
\end{matrix}
\right)  =0.
\end{aligned}
\end{equation}
Consequently, its Lagrangian function reads as:
\begin{equation*}
\begin{aligned}
L(u,v,x,z,\lambda_1, \lambda_2)=&\tau\Vert|x|\Vert_1+\mu\Vert v\Vert_*+\rho_{c}(z-b_0)\\
&+\langle\lambda_1,\nabla u-x\rangle+\langle\lambda_2,\Phi(u+v)-z\rangle .  
\end{aligned}
\end{equation*}
Clearly, due to convexity of \eqref{dual_pro}, it easily follows from  \cite[Section 5.3.2]{BV03} that solving \eqref{dual_pro} is equivalent to finding a saddle point of the following problem:
\begin{equation}\label{saddle}
\min_{(u,v,x,z)}\max_{(\lambda_1,\lambda_2)}L(u,v,x,z,\lambda_1,\lambda_2).
\end{equation}
To further simplify \eqref{saddle} and reformulate it as a structured saddle point problem, by using the notation $\bm{x}=(u^\top,v^\top)^\top$, $\bm{y}=(x^\top, z^\top)^\top$, $\bm{\lambda}=(\lambda_1^\top, \lambda_2^\top)^\top$, it follows from the definition of conjugate function and \eqref{huber_con} that
$$
\begin{aligned}
\min_{\bm{y}}L(\bm{y}, \bm\lambda)&=\min_{x, z} \tau\Vert |x| \Vert_1 + \rho_{c}(z-b_0) - \langle\lambda_1,x\rangle - \langle\lambda_2,z\rangle\\
&= \sup_{x, z} \langle\lambda_1,x\rangle + \langle\lambda_2,z\rangle - \tau\Vert |x| \Vert_1 - \rho_{c}(z-b_0)\\
& = \sup_{x}\{\langle\lambda_1,x\rangle - \tau\Vert |x| \Vert_1\} \\
&\quad + \sup_{z}\{\langle\lambda_2,z\rangle - \rho_{c}(z-b_0)\}\\
& = (\tau\Vert |x| \Vert_1)^*(\lambda_1) + (\rho_{c}(z-b_0))^*(\lambda_2)\\
& = \delta_{\Vert \lambda_1 \Vert_{\infty}\leq \tau} + \rho^*_{c}(\lambda_2)+\langle\lambda_2, b_0\rangle\\
& = F^*(\lambda_1, \lambda_2),
\end{aligned}
$$
which immediately implies that \eqref{saddle} can be recast as
\begin{equation*}
\begin{aligned}
\min_{u,v} \max_{\lambda_1,\lambda_2}\
&\underbrace{\langle\nabla u, \lambda_1\rangle+\langle\Phi(u + v),\lambda_2\rangle}_{\langle K\bm{x},\bm{\lambda}\rangle} +\underbrace{\mu\Vert v\Vert_*}_{G(\bm{x})}
\\
&-\underbrace{\left(\delta_{\Vert\lambda_1\Vert_{\infty}\leq\tau}+\rho^*_{c}(\lambda_2)+\langle\lambda_2,b_0\rangle\right)}_{F^*(\bm{\lambda})\equiv F^*(\lambda_1,\lambda_2)} ,
\end{aligned}
\end{equation*}
where 
\begin{equation*}
K\bm{x}= \left(\begin{matrix}
\nabla u \\
\Phi(u+v)\\
\end{matrix}\right) = \left(\begin{matrix}
\nabla  & O \\
\Phi & \Phi\\
\end{matrix}\right) \left(\begin{matrix}
u \\
v\\
\end{matrix}\right).
\end{equation*}
Therefore, by the employment of the first-order primal-dual algorithm \cite{CP11} for 
\begin{equation}\label{c-p_problem}
\min_{\bm{x}} \max_{{\bm \lambda}}\;\;\langle K\bm{x},\bm{\lambda}\rangle+G(\bm{x})-F^*(\bm{\lambda}),
\end{equation}
the iterative scheme is specified as
\begin{equation}\label{c-p_iter}
\left\{
\begin{aligned}
&\bm{\lambda}^{k+1}=\text{prox}_{\sigma F^*}(\bm{\lambda}^k+\sigma K \bar{\bm{x}}^k),\\
& \bm{x}^{k+1}=\text{prox}_{\eta G}(\bm{x}^k-\eta K^\top\bm{\lambda}^{k+1}),\\
&\bar{\bm{x}}^{k+1}=2\bm{x}^{k+1}-\bm{x}^k.
\end{aligned}\right.
\end{equation}
More concretely, for given the $k$-th iteration, $\lambda_1^k$, $\lambda_2^k$, $u^k$, $v^k$, $\bar{u}^k$, and $\bar{v}^k$, the iterative scheme \eqref{c-p_iter} is specified as
\begin{itemize}
	\item Update dual variables $\lambda_1^{k+1}$ and $\lambda_2^{k+1}$ via:
	\begin{align}
	\lambda_1^{k+1} &= \text{prox}_{\sigma\delta_{\Vert \lambda_1 \Vert_{\infty}\leq\tau}}(\lambda^k_1+\sigma\nabla\bar{u}^k) \nonumber \\
	&= \text{Clip}_{\tau}(\lambda^k_1+\sigma\nabla\bar{u}^k), \label{dual_update-1}\\
	\lambda_2^{k+1} &= \text{prox}_{\sigma(\rho^*_{c}(\lambda_2)+\langle \lambda_2, b_0\rangle)}(\lambda_2^k+\sigma \Phi(\bar{u}^k + \bar{{v}^k})) \nonumber \\
	&= \text{prox}_{\sigma\rho^*_{c}(\cdot)}(\lambda_2^k+\sigma \Phi(\bar{u}^k + \bar{v}^k)-\sigma b_0), \label{dual_update-2}
	\end{align}
	where the proximity operator of the Huber function's conjugate function is defined as:
	\begin{equation*}\label{huber_con_prox}
	\text{prox}_{\sigma \rho_c^*}(x) = \left\{
	\begin{array}{ll}
	\frac{x}{1+\sigma},  & \text{if $|x|\leq (\sigma+1)c$}, \\
	c\cdot\text{sign}(x) , & \text{if $|x|> (\sigma+1)c$}.
	\end{array}
	\right.
	\end{equation*}
	\item Update primal variables $u^{k+1}$ and $v^{k+1}$ via:
	\begin{align}
	u^{k+1}&=u^k-\eta\left(\nabla^\top\lambda_1^{k+1}+\Phi^\top\lambda_2^{k+1}\right),\label{primal_update-1}\\
	v^{k+1} &= \text{prox}_{\eta\mu\Vert\cdot\Vert_*}(v^k-\eta\Phi^\top\lambda_2^{k+1}) \nonumber \\
	&= \mathcal{S}\left(v^k-\eta\Phi^\top\lambda_2^{k+1}, \eta\mu\right). \label{primal_update-2}
	\end{align}
	\item Do the extrapolation step:
	\begin{align}
	\bar{u}^{k+1} &= 2 u^{k+1} - u^k,\label{middel_update-1}\\
	\bar{v}^{k+1} &= 2 v^{k+1} - v^k. \label{middel_update-2}
	\end{align}
\end{itemize}

With the above preparations, we can summarize them in Algorithm \ref{alg_II}.

\renewcommand{\algorithmicrequire}{\textbf{Require:}}
\begin{algorithm}[!htbp]
	\setstretch{1.1} 
	\caption{Solving~problem~\eqref{dual_pro} with $\Phi\neq I$.} \label{alg_II}
	\begin{algorithmic}[1]
		\REQUIRE Set initial points $(u^0,v^0,\lambda_1^0,\lambda_2^0, \bar{u}^0,\bar{v}^0)$, and $\sigma>0$, $\eta>0$ satisfying $\sqrt{\sigma\eta}< \frac{1}{\|K\|}$.
		\FOR{$k=0,1,\cdots$}
		\STATE Update $\lambda_{1}^{k+1}$ and $\lambda_{1}^{k+1}$ via \eqref{dual_update-1} and \eqref{dual_update-1}, respectively.
		\STATE Update $u^{k+1}$ and $v^{k+1}$ via \eqref{primal_update-1} and \eqref{primal_update-1}, respectively.
		\STATE Compute $\bar{u}^{k+1}$ and  $\bar{v}^{k+1}$ via \eqref{middel_update-1} and \eqref{middel_update-2}, respectively.
		\ENDFOR
	\end{algorithmic}
\end{algorithm}

Under the condition $\sqrt{\sigma\eta}<\frac{1}{\| K\|}$, it follows from \cite[Theorem 1]{CP11} that Algorithm \ref{alg_II} also enjoys a global convergence and the $\mathcal{O}(1/N)$ convergence rate. The complete proof is referred to \cite{CP11} for the conciseness of this paper.

\begin{theorem}\label{theorem_2}
	Let $(\hat{\bm{x}}, \hat{\bm{\lambda}})$ be a saddle point of  \eqref{c-p_problem} and $\{(\bm{x}^k, \bm{\lambda}^k, \bar{\bm{x}}^k)\}$ be the sequence generated by \eqref{c-p_iter} (i.e., Algorithm \ref{alg_II}). Then:
	\begin{enumerate}
		\item[(a).]  There exists a saddle-point $(\bm{x}^\star,\bm{\lambda}^\star)$ such that $\bm{x}^k\to \bm{x}^\star$ and $\bm{\lambda}^k\to \bm{\lambda}^\star$.
		\item[(b).] For any $k$, $\{(\bm{x}^k,\bm{\lambda}^k)\}$ remains bounded and satisfies
		\begin{align*}
		&\frac{\|\bm{\lambda}^k-\hat{\bm{\lambda}}\|^2}{2\sigma}+\frac{\|\bm{x}^k-\hat{\bm{x}}\|^2}{2\eta}\\
		&\qquad \leq C\left(\frac{\|\bm{\lambda}^0-\hat{\bm{\lambda}}\|^2}{2\sigma}+\frac{\|\bm{x}^0-\hat{\bm{x}}\|^2}{2\eta}\right),
		\end{align*}
		where the constant $C\leq (1-\eta\sigma \|K\|^2)^{-1}$.
		\item[(c).] Let $\bm{x}^N=(\sum_{k=1}^N \bm{x}^k)/N$ and $\bm{\lambda}^N=(\sum_{k=1}^N \bm{\lambda}^k)/N$, for any bounded set $\mathbb{B}_1\times \mathbb{B}_2$, the restricted primal-dual gap has the following bound:
		\begin{equation}
		\mathcal{G}_{\mathbb{B}_1\times \mathbb{B}_2}(\bm{x}^N, \bm{\lambda}^N)\leq\frac{\mathcal{D}(\mathbb{B}_1,\mathbb{B}_2)}{N},
		\end{equation}
		where 
		\begin{align*}
		\mathcal{G}_{\mathbb{B}_1\times \mathbb{B}_2}(\bm{x}, \bm{\lambda})=&\max_{\bm{\lambda}'\in \mathbb{B}_2} \left\{\langle \bm{\lambda}', K \bm{x}\rangle-F^*(\bm{\lambda}')+G(\bm{x})\right\}\\
		&-\min_{\bm{x}'\in \mathbb{B}_1} \left\{\langle \bm{\lambda}, K \bm{x}'\rangle-F^*(\bm{\lambda})+G(\bm{x}')\right\},
		\end{align*}
	 represents the primal-dual gap and
		$$ \mathcal{D}(\mathbb{B}_1,\mathbb{B}_2) = \sup_{(\bm{x},\bm{\lambda})\in \mathbb{B}_1\times \mathbb{B}_2} \left\{\frac{\|\bm{x}-\bm{x}^0\|^2}{2\eta}+\frac{\|\bm{\lambda}-\bm{\lambda}^0\|^2}{2\sigma}\right\}. $$
		Moreover, the weak cluster points of $(\bm{x}^N, \bm{\lambda}^N)$ are saddle-points of \eqref{c-p_problem}.		
	\end{enumerate}
\end{theorem}

\section{Experimental Results}\label{sec_exp}

In this section, we conduct the performance of our RLRP model \eqref{RLRP} on some image processing problems. We consider three scenarios in our experiments: (i). $\Phi = I$, which corresponds to the pure image denoising of an observation with heavy-tailed noise. (ii). $\Phi = S$, which represents a down-sampling operator including the binary ``mask'' and random sampling. (iii). $\Phi = B$, which is a blurring operator. To verify the robustness of our model to heavy-tailed noise, for all the above cases, we will compare our approach (denoted by \textbf{RLRP}) with the G-norm model \cite{NYZ13} (\textbf{NYZ} in short), the LPR model \cite{HKZ15} (\textbf{HKZ} in short), the BNN model  \cite{OMY14} (\textbf{OMY} in short), and the customized LRP model \cite{ZH21} (\textbf{CLRP} in short). In addition, there are recent models such as the $L_0$ gradient model \cite{PWH24} (\textbf{PWH} in short), the WLS model \cite{LWC25} (\textbf{LWC} in short), and the PnP network model \cite{GAT25} (\textbf{GAT} in short).

We set the initial points as $u^0=\bar{u}^0=b_0$, and other variable are taken as zero matrices (or vectors). Besides, the stopping criterion is taken as follows for all experiments:
\begin{equation}\label{stopping_criterion}
\mathrm{Tol} = \max\left\{\frac{\norm{u^{k+1}-u^k}}{\norm{u^k}+1},\frac{\norm{v^{k+1}-v^k}}{\norm{v^k}+1}\right\}<\epsilon,
\end{equation}
where the tolerance $\epsilon$ is a positive constant, which was set the same value of $\epsilon$ in the same experiment.

\begin{figure*}[!t]
	\centering
	\includegraphics[width = 0.85\textwidth]{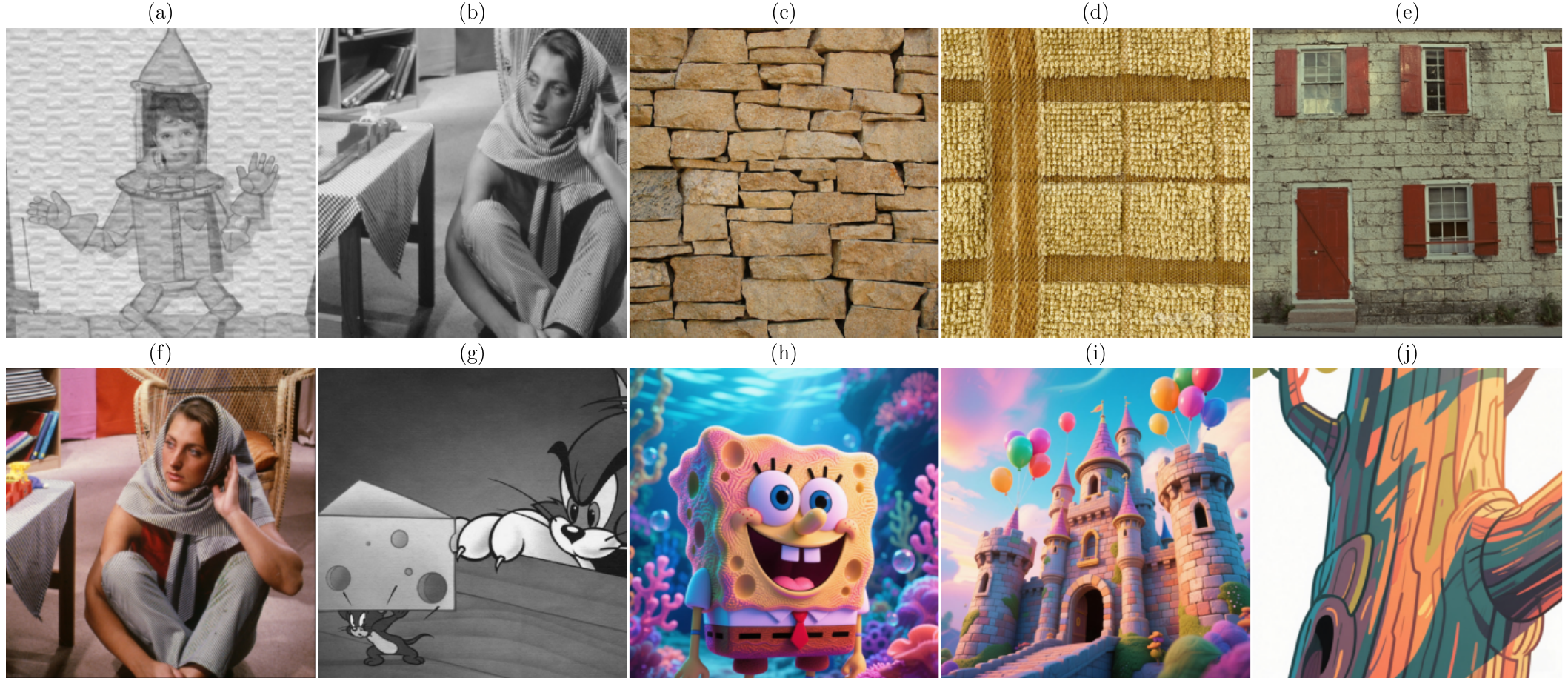}
	\caption{Test images. (a) \textsf{Boy} synthetic image ($256 \times 256$); (b) \textsf{Barbara} natural image ($512 \times 512$); (c)  \textsf{Stone} natural image ($512\times 512 \times 3$); (d) \textsf{Towel} natural image ($512\times 512 \times 3$); (e)  \textsf{Kodim\_wall} natural image ($512\times 512\times 3$);
		(f) \textsf{Barbara\_RGB} natural image ( $512 \times 512 \times 3$ ); (g) \textsf{TomJerry} cartoon image ($512 \times 512$); (h)  \textsf{Sponge\_Bob} cartoon image ($512 \times 512 \times 3$); (i)  \textsf{Castle} cartoon image ($512 \times 512 \times 3$); (j)  \textsf{Bole} cartoon image ($512 \times 512 \times 3)$.}
	\label{select_fig}
\end{figure*}

All test images are displayed in Fig. \ref{select_fig}. 
Among the selected images, there are grayscale images (a), (b) and RGB color images (c)-(j). Image (a) is a synthetic image, formed by superimposing the cartoon component and the texture component in a 7:3 ratio. Images (b)-(f) are real-world images and have relatively distinct and well-preserved texture patterns. Images (g)-(j) are colored piece-wise continuous and smooth cartoon-like images with distinct locally clear textures. And all tested images are scaled into $[0,1]$. All numerical experiments are implemented in {\sc Python} 3.11 and are conducted on a laptop computer with Intel(R) Core(TM) i5 CPU 2.61GHz and 16G memory.

\subsection{Pure image denoising}\label{subsec:case-I}

In this subsection, we consider the pure image denoising of observed images corrupted by heavy-tailed noise. We select $8$ images and divided into $2$ groups. These images are used to test the restoration ability of the model for grayscale images, RGB color images, synthetic images, real-world images, cartoon-like images and images with good patterns respectively.

Regarding the parameter settings, for the synthetic image (a), we set $\tau = 0.1$ and $\mu = 2$ for model \eqref{RLRP}. For the rest of images, we set $\tau = 0.015$ and $\mu = 0.2$. For these algorithmic parameters, as described in \cite{ZH21} for \eqref{CLRP}, we set $\gamma = 1.3$ for image (a), while for other images, we set $\gamma = 1.6$, and $r=1, s=2.01$ for all.
The student's-t distribution is a typical symmetric heavy-tailed distribution. As the degrees of freedom increase, it increasingly approaches the Gaussian distribution. When the degrees of freedom is $2$, its heavy-tailed effect is the most pronounced (see \cite{t-distribution}).
For gray images (a), (b), (g), we added standard Student's-t noise with intensity of $0.1$ and $0.2$ for rest. The degrees of freedom are both $2$. For the synthetic image (a), we set the threshold parameter $c=0.1$ of the Huber function, the algorithmic parameter $\beta=2$, and the stopping value $\epsilon=10^{-2}$ given in \eqref{stopping_criterion}. For the other images, we set $\beta = 0.2$, $\epsilon = 10^{-2}$ and the threshold $c = 0.01$. The signal-to-noise ratio (SNR), which is typically used to evaluate the quality of the image, is defined by
$$\mathrm{SNR} = 20\log_{10} \frac{\norm{b}}{\norm{\tilde{b}-b}},$$
where $b$ is the original image, $\tilde{b}$ is the restored image. Another popular metric used to evaluate image quality is the \textit{structural similarity} (SSIM, see details in \cite{ssim}). 

\begin{figure*}[!t]
	\centering
	\includegraphics[width = 0.95\textwidth]{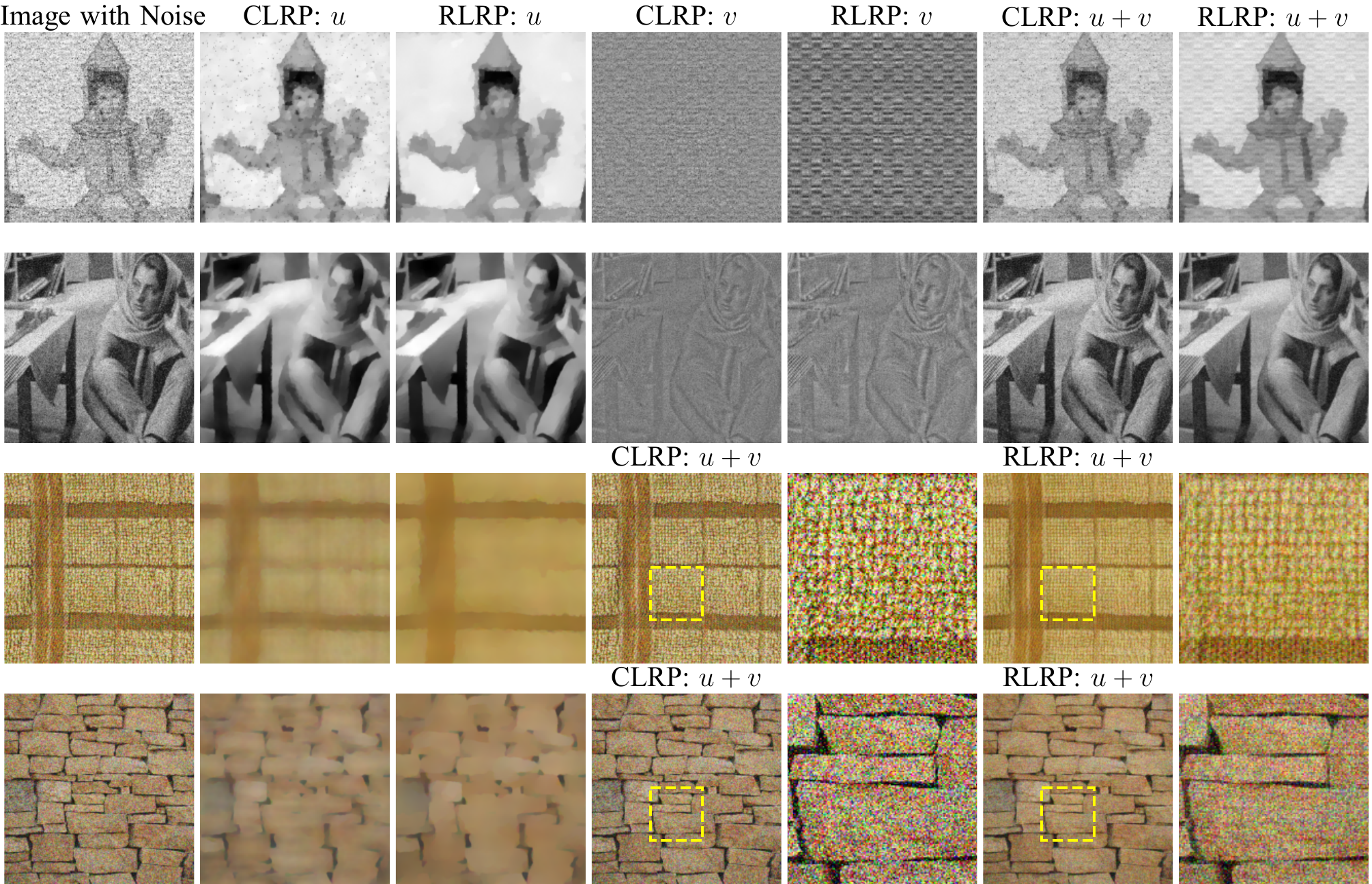}
	\caption{Results for clear image decomposition: $\Phi=\bm{I}$. From left to right: the noisy image, comparison of the cartoon part, comparison of the texture part, comparison of the restored images (left: CLRP, right: RLRP). From top to bottom: \textsf{Boy}, \textsf{Barbara}, \textsf{Towel} and \textsf{Stone} respectively.}
	\label{I1}
\end{figure*}

\begin{figure*}[!t]
	\centering
	\includegraphics[width = 0.95	\textwidth]{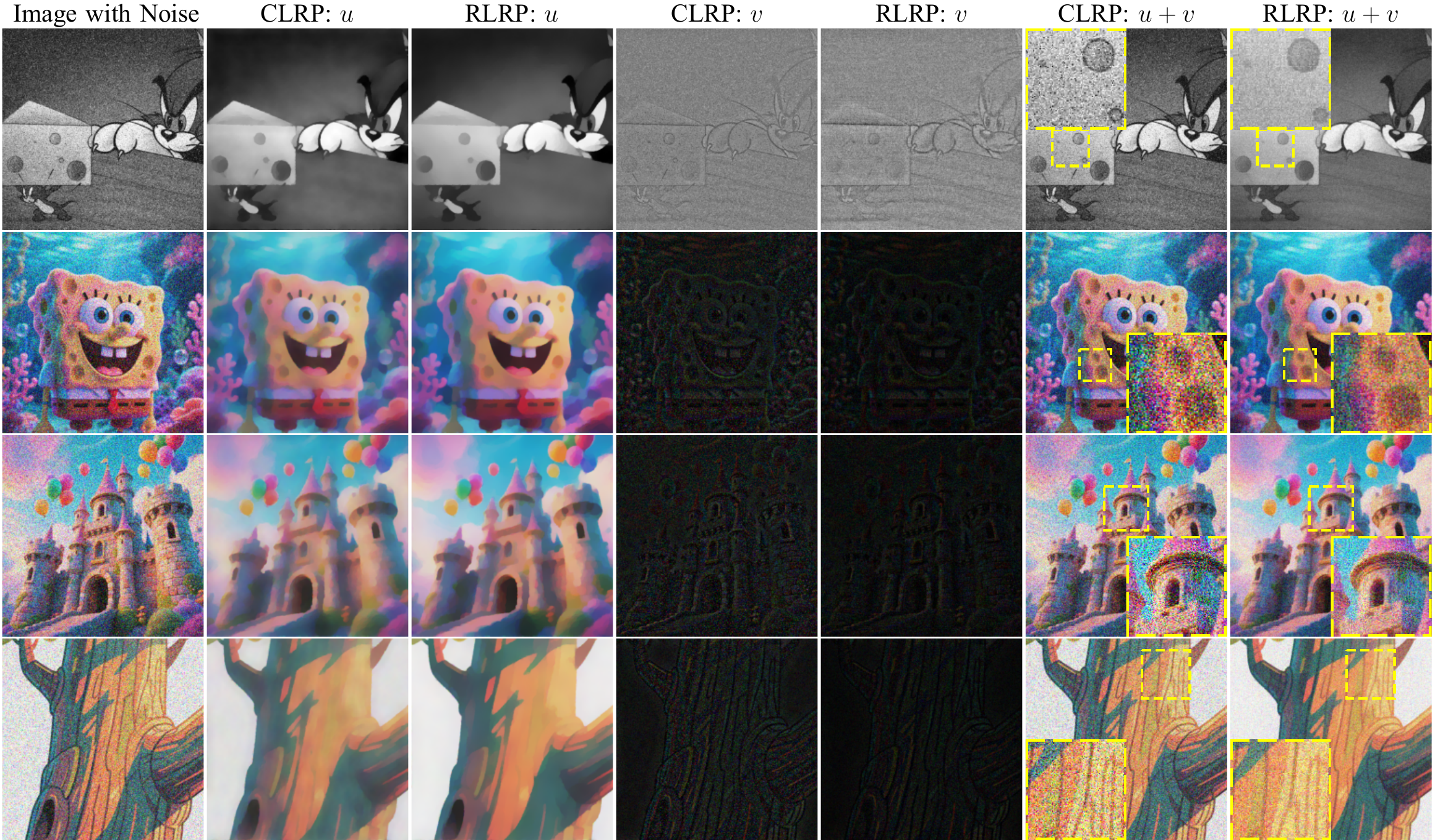}
	\caption{Results for cartoon image decomposition: $\Phi=\bm{I}$. From left to right: the noisy image, comparison of the cartoon part, comparison of the texture part, comparison of the restored images (left: CLRP, right: RLRP). From top to bottom: \textsf{TomJerry}, \textsf{Sponge\_Bob}, \textsf{Castle} and \textsf{Bole} respectively.}
	\label{I2}
\end{figure*}

\begin{figure*}[!t]
	\centering
	\includegraphics[width = 0.95	\textwidth]{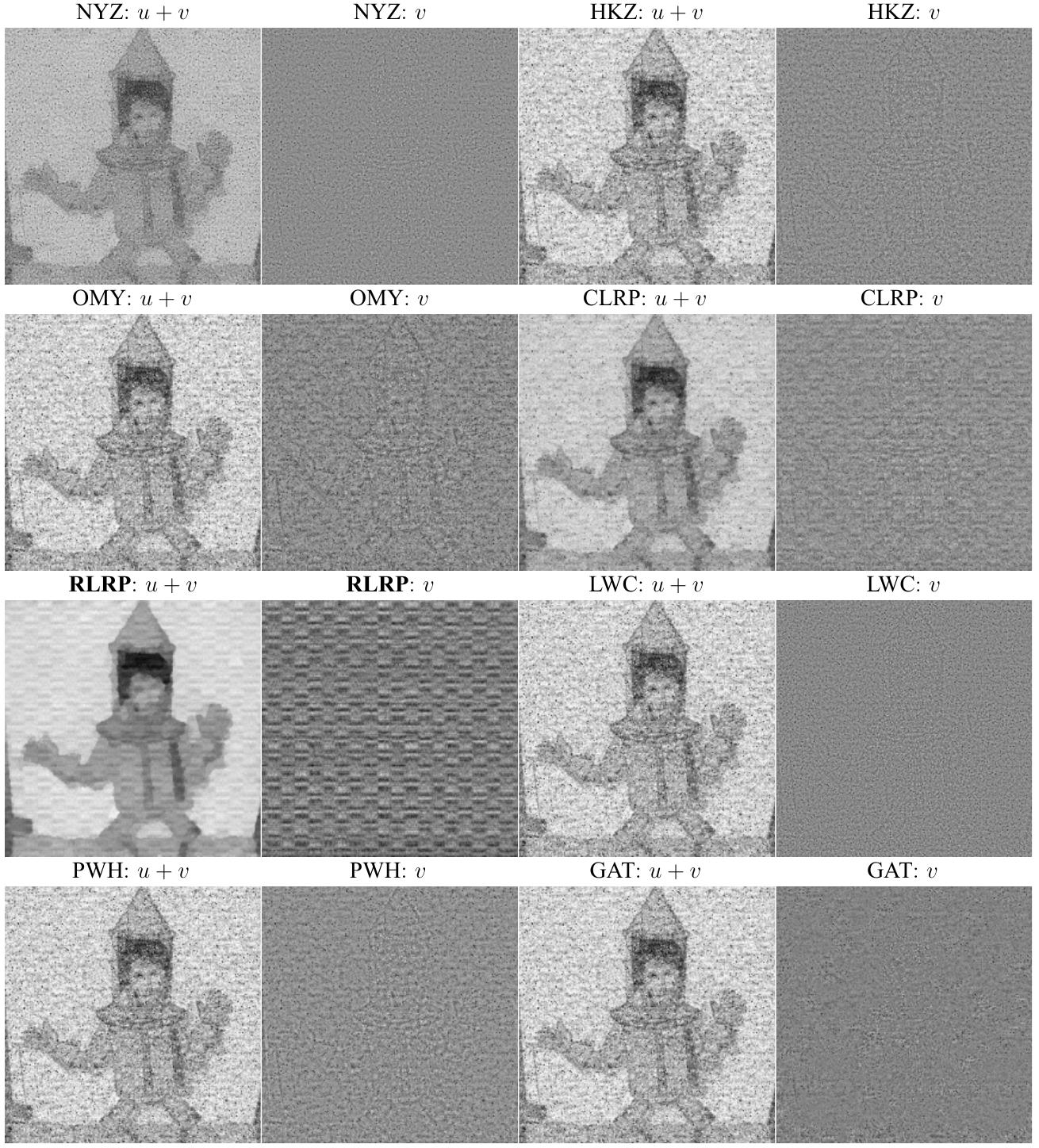}
	\caption{The decomposition results of \textsf{Boy} under different models (left: $u+v$, right: $v$.)}
	\label{compare_I}
\end{figure*}

\begin{figure}[!htbp]
	\centering
	\includegraphics[width = 0.225\textwidth]{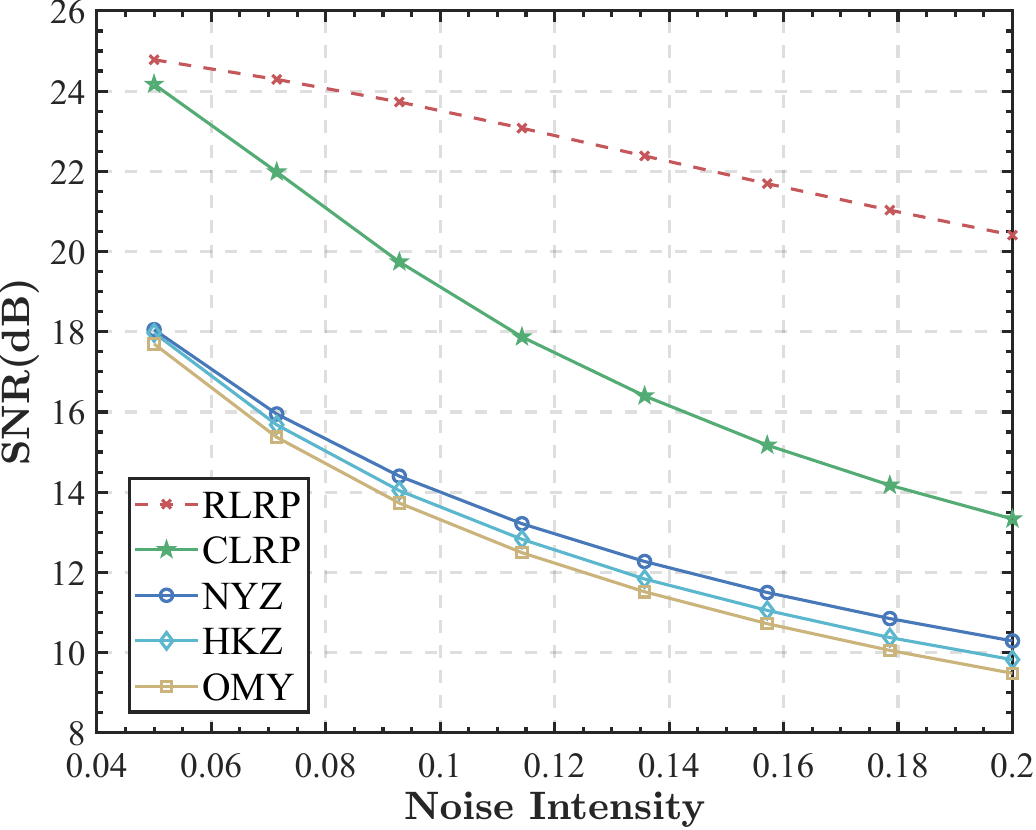}
	\includegraphics[width = 0.225\textwidth]{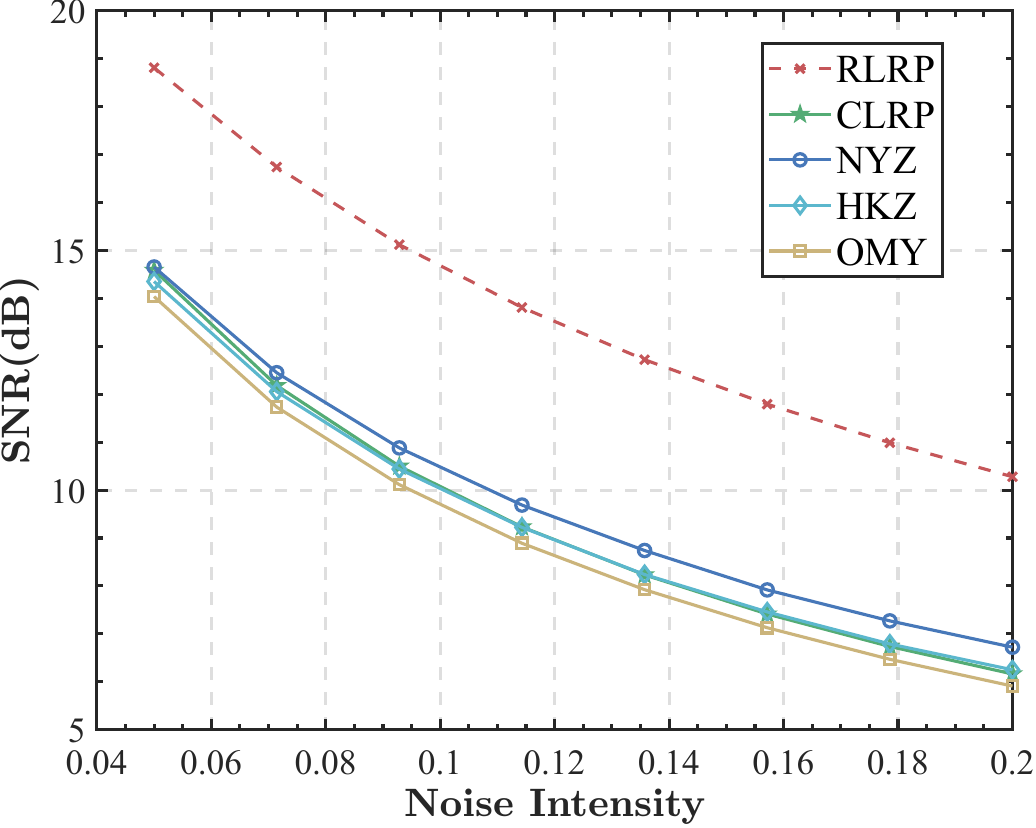}
	\caption{The SNR values of different noise intensity. Left: \textsf{Boy}, right: \textsf{Barbara}.}
	\label{SNR_analysis}
\end{figure}

Based on the numerical results, the performance levels of the existing models are comparatively similar. To enhance clarity and conciseness, rather than enumerating the results of every model for all images individually, we have chosen the CLRP model as a representative benchmark. Its comparative performance against the RLRP model is illustrated in Fig. \ref{I1} and Fig. \ref{I2}. The complete set of numerical results is provided in Table \ref{result_I}.

As illustrated in Fig. \ref{I1}, the CLRP model struggles to identify outliers under significant heavy-tailed noise interference. Consequently, the cartoon component is heavily corrupted by outliers. This is particularly evident in the synthetic image, where numerous noise points remain unsmoothed. Additionally, the texture component is also adversely affected, leading to the misclassification of certain cartoon patterns as texture. For natural images, CLRP and RLRP exhibit broadly similar performance due to their analogous algorithmic structures. 
However, in the restoration results of cartoon-like images in Fig. \ref{I2}, RLRP shows better textures and less noise.
We attribute this difference to the disruptive effects of high-intensity heavy-tailed noise, which compromises the inherent low-rank properties of certain patterns. 
Compared with CLRP, the adaptive thresholding mechanism of RLRP can effectively identify outliers and utilize prior information to restore latent textures.
To further evaluate noise resilience across models, we compared the restoration performance of different models. The experimental results show that, apart from RLRP which can achieve excellent denoising results, the performance of other models is similar to that of CLRP, lacking a mechanism for outlier detection. Figure \ref{compare_I} presents the restoration results of image (a) obtained by all comparative models, including the restored images and the reconstructed texture component $v$. Experimental results show that the overall smoothness of the image restored by the RLRP model is significantly better than that of other models, with no obvious noise remaining, and the extracted texture component is fine and realistic. Other models suffer from strong noise interference, losing accurate texture discrimination ability; their extracted texture components are covered or even completely corrupted by noise. These results verify that the proposed model has excellent resistance to heavy-tailed noise.

Furthermore, we conduct a comprehensive evaluation of noise robustness by comparing the proposed model with existing approaches across varying noise intensities. As demonstrated in Fig. \ref{SNR_analysis}, our model consistently outperforms the other methods, maintaining superior performance under both mild and severe noise conditions. This robust performance advantage is particularly evident in the signal-to-noise ratio analysis.

\begin{figure}[!htbp]
	\centering
	\includegraphics[width = 0.225\textwidth]{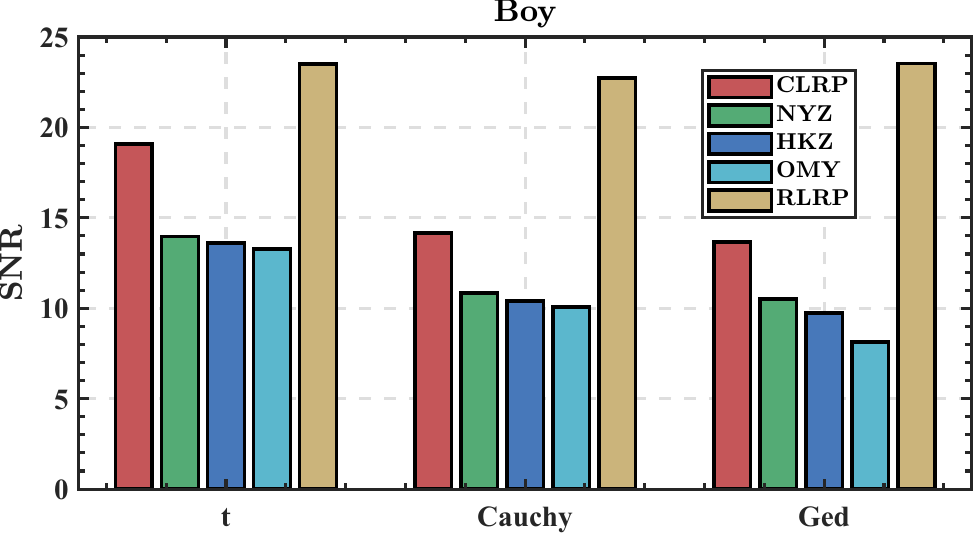}
	\includegraphics[width = 0.225\textwidth]{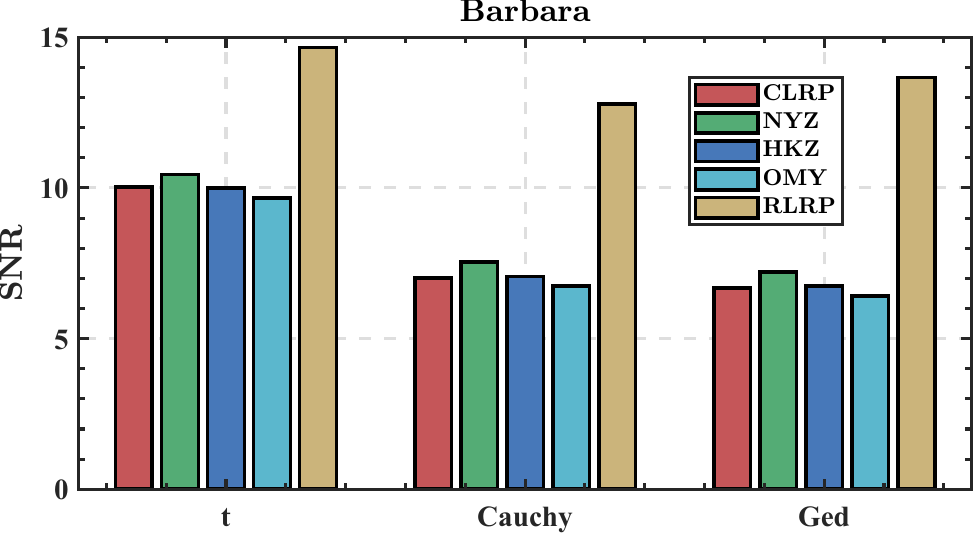}
	\includegraphics[width = 0.225\textwidth]{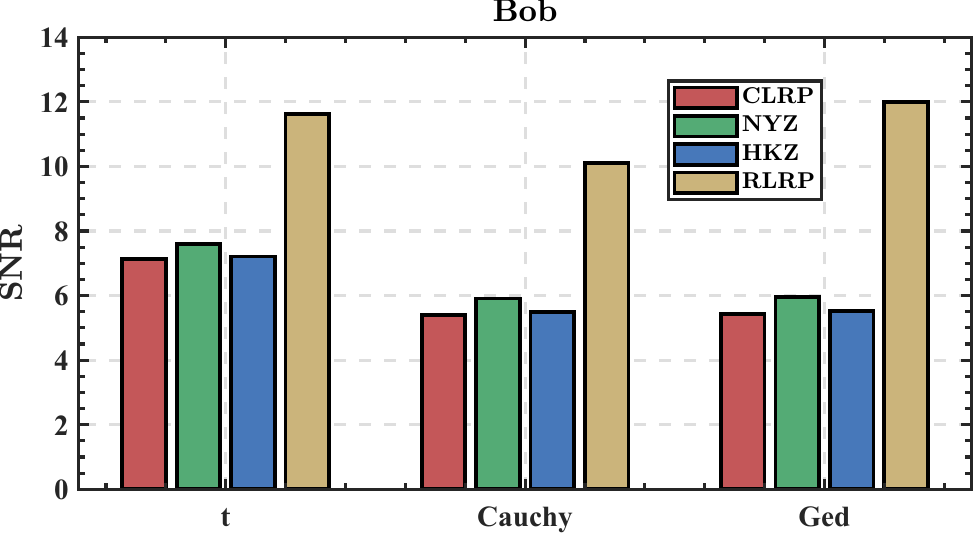}
	\includegraphics[width = 0.225\textwidth]{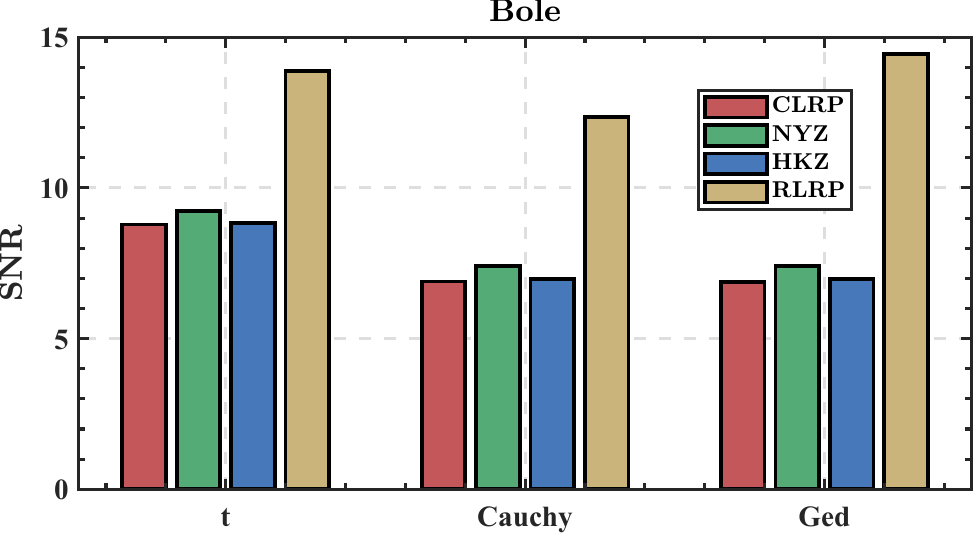}
	\caption{The SNR values of images restored by different models under different types of noise, in the order of \textsf{Boy}, \textsf{Barbara}, \textsf{Sponge\_Bob}, \textsf{Bole}.}
	\label{noise_analysis}
\end{figure}

To rigorously validate the robustness of our proposed model, we conducted extensive experiments with multiple heavy-tailed noise distributions beyond the t-distribution, including the Cauchy distribution and Generalized Error Distribution (GED).

The Cauchy distribution represents an extreme case of heavy-tailed behavior. Unlike distributions with finite moments, it lacks well-defined mean and variance, with its tail decaying at a rate of $x^{-2}$. This characteristic leads to frequent extreme outliers, making it significantly heavier-tailed than the t-distribution (see \cite{Arnold79}). The GED offers more flexible tail behavior through its shape parameter. While maintaining heavy-tailed properties, its tail decay is moderately lighter than the t-distribution (see \cite{BT92}).

As evidenced by Fig. \ref{noise_analysis}, our RLRP model demonstrates exceptional resilience against all tested heavy-tailed noises, consistently maintaining superior SNR values compared to baseline methods.

All numerical results are summarized in Table \ref{result_I}, with the proposed approach's performance highlighted in bold for clarity. Our approach demonstrates consistently superior performance across nearly all test images, achieving SNR values that are 3-5 dB higher than existing methods. This significant improvement clearly validates the approach's enhanced capability in handling high-intensity noise interference. Regarding computational efficiency, while our model shares the same regularization terms with the CLRP, its more difficult data fidelity term leads to increased solution complexity. Consequently, the computational time is approximately doubled compared to baseline algorithms. In future, we will discuss potential optimization strategies to achieve a better trade-off between performance gains and computational overhead for practical applications.

\begin{table}[h]
	\centering
	\caption{Restoration results for the case $\Phi = \bm{I}$ (with t-noise).}
	\resizebox{\linewidth}{!}{
		\begin{tabular}{c|ccccc}
			\hline
			\cline{2-6}
			Image & Method & SNR(dB) & Iter. & Time(s) & SSIM \\
			\hline
			\multirow{5}{*}{\textsf{Boy}} 
			& CLRP & 19.071 &  11 & 0.23 & 0.433 \\
			& NYZ & 13.971 & 28 & 0.30 & 0.219\\
			& HKZ & 13.604 & 25 & 0.22 & 0.260\\
			& OMY & 13.278 & 32 & 1.14 & 0.253 \\
            & LWC & 13.279 & 6 & 1.02 & 0.252 \\
            & PWH & 13.279 & 41 & 0.21 & 0.253 \\
            & GAT & 13.269 & 10 & 4.11 & 0.253 \\
			& \textbf{RLRP} & \textbf{23.520} &  36 & 0.64 & \textbf{0.603} \\
			\hline
			\multirow{5}{*}{\textsf{Barbara}} 
			& CLRP & 10.038 &  17 & 1.63 & 0.320 \\
			& NYZ & 10.448 & 27 & 1.53 & 0.320\\
			& HKZ & 9.990 & 25 & 1.08 & 0.318\\
			& OMY & 9.670 & 44 & 13.82 & 0.310 \\
            & LWC & 9.669 & 7 & 7.028 & 0.309 \\
            & PWH & 9.669 & 44 & 1.19 & 0.310 \\
            & GAT & 9.671 & 10 & 8.92 & 0.310 \\
			& \textbf{RLRP} & \textbf{14.663} &  32 & 2.68 & \textbf{0.465} \\
			\hline
			\multirow{4}{*}{\textsf{Towel}} 
			& CLRP & 7.292  &  20 & 6.15 & 0.555 \\
			& NYZ & 7.331 & 28 & 5.53 & 0.444\\
			& HKZ & 7.357 & 29 & 4.21 & 0.553\\
            & LWC & 10.776 & 6 & 6.740 & 0.563 \\
            & PWH & 10.750 & 41 & 1.06 & 0.597 \\
            & GAT & 10.740 & 10 & 8.92 & 0.596 \\
			& \textbf{RLRP} & \textbf{11.471} &  46 & 14.65 & \textbf{0.571} \\
			\hline
			\multirow{4}{*}{\textsf{Stone}} 
			& CLRP & 6.240  &  19 & 5.46 & 0.327 \\
			& NYZ & 6.695 & 26 & 4.97 & 0.284\\
			& HKZ & 6.338 & 29 & 4.28 & 0.327\\
            & LWC & 9.674 & 6 & 8.36 & 0.340 \\
            & PWH & 9.737 & 44 & 1.14 & 0.356 \\
            & GAT & 9.730 & 10 & 8.87 & 0.356 \\
			& \textbf{RLRP} & \textbf{9.985} &  39 & 9.98 & \textbf{0.419} \\
			\hline
			\multirow{4}{*}{\textsf{TomJerry}} 
			& CLRP & 9.335 &  17 & 1.94 & 0.182 \\
			& NYZ & 9.787 & 27 & 1.87 & 0.189\\
			& HKZ & 9.281 & 25 & 1.24 & 0.181\\
			& OMY & 8.945 & 43 & 18.26 & 0.176\\
            & LWC & 12.396 & 6 & 7.43 & 0.254 \\
            & PWH & 12.396 & 50 & 1.29 & 0.254 \\
            & GAT & 12.393 & 10 & 8.96 & 0.254 \\
			& \textbf{RLRP} & \textbf{16.317} &  38 & 3.51 & \textbf{0.385} \\
			\hline
			\multirow{4}{*}{\textsf{Sponge\_Bob}} 
			& CLRP & 7.143 &  17 & 5.68 & 0.212 \\
			& NYZ & 7.598 & 26 & 5.39 & 0.191\\
			& HKZ & 7.213 & 28 & 4.08 & 0.213\\
            & LWC & 9.495 & 6 & 7.07 & 0.195 \\
            & PWH & 9.794 & 47 & 1.24 & 0.236 \\
            & GAT & 9.776 & 10 & 8.84 & 0.236 \\
			& \textbf{RLRP} & \textbf{11.623} &  37 & 11.36 & \textbf{0.333} \\
			\hline
			\multirow{4}{*}{\textsf{Castle}} 
			& CLRP & 7.965 &  17 & 5.29 & 0.184 \\
			& NYZ & 8.448 & 26 & 5.02 & 0.169\\
			& HKZ & 8.033 & 28 & 3.89 & 0.185\\
            & LWC & 10.648 & 6 & 6.96 & 0.193 \\
            & PWH & 10.718 & 47 & 1.24 & 0.236 \\
            & GAT & 10.699 & 10 & 8.81 & 0.207 \\
			& \textbf{RLRP} & \textbf{12.416} &  38 & 10.72 & \textbf{0.294} \\
			\hline
			\multirow{4}{*}{\textsf{Bole}} 
			& CLRP & 8.789 &  17 & 5.09 & 0.199 \\
			& NYZ & 9.231 & 26 & 5.06 & 0.170\\
			& HKZ & 8.842 & 28 & 4.43 & 0.200\\
            & LWC & 11.625 & 6 & 7.31 & 0.203 \\
            & PWH & 11.626 & 47 & 1.31 & 0.227 \\
            & GAT & 11.599 & 10 & 8.85 & 0.227 \\
			& \textbf{RLRP} & \textbf{13.882} &  36 & 10.02 & \textbf{0.335} \\
			\hline
	\end{tabular}}
	\label{result_I}
\end{table}

\begin{figure*}[!t]
	\centering
	\includegraphics[width = 0.85\textwidth]{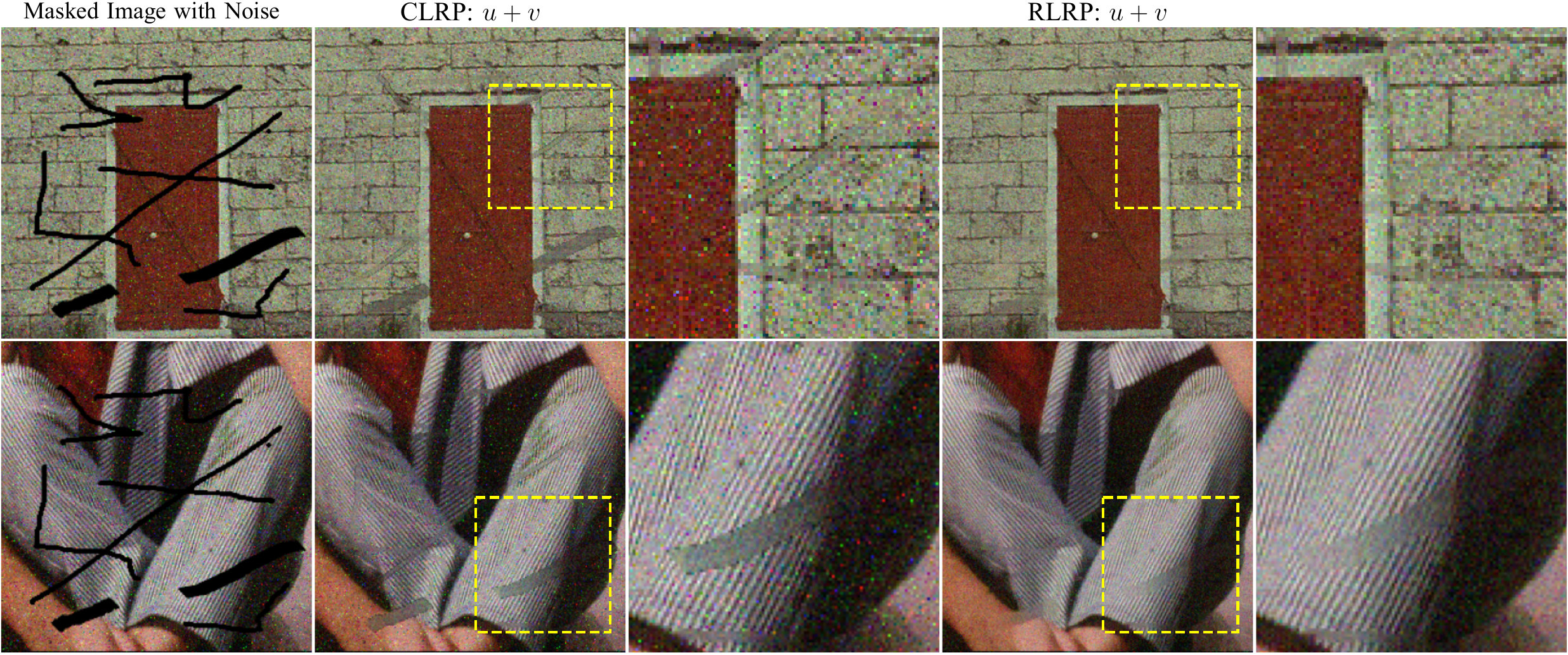}
	\caption{Results for masked images: $\Phi=\bm{S}$. From top to bottom: \textsf{Kodim\_wall}, \textsf{Barbara\_RGB}. From left to right: the masked images, comparison of the restored images, comparison of the locally magnified images. (left CLRP, right RLRP).}\label{mask}
\end{figure*}

\subsection{Image inpainting}
In this part, we are concerned the performance of our approach on image inpainting. 

First, for the mask down-sampling operator, we select three images to conduct the noise resistance ability of the model for low-rank patterns, grayscale images, and color images respectively.
Specifically, we add t-noise with an intensity of $0.05$ to the original images, and then add a binary mask. 

Regarding the model parameters, we uniformly set $\tau = 0.015$, $\mu = 0.2$. For the algorithm of solving \eqref{CLRP} introduced in \cite{ZH21}, we set $\beta = 2$, $\gamma = 1.6$, $r = 1$, $s = 2.01$. For our approach, we uniformly set $\sigma=\eta = 0.35$ and $500$ as the maximum number of iterations. For each image, the stopping tolerance is set to $\epsilon = 2.5\times10^{-4}$. 

Fig. \ref{mask} clearly demonstrates the superior performance of our RLRP compared to the original CLRP when using identical stopping criteria. Our approach exhibits several notable advantages in texture reconstruction: (i) effective noise suppression while maintaining texture fidelity, (ii) precise restoration of missing components through rigorous low-rank optimization rather than simple pixel approximation, and (iii) significantly cleaner visual outputs that highlight its enhanced noise resistance capabilities relative to the CLRP model.

These qualitative observations are quantitatively supported by the results presented in Table \ref{result_mask}, which shows that our RLRP  consistently achieves SNR improvements of 2-3 dB across all test images compared to other approaches. The combination of superior visual quality and measurable performance metrics confirms the effectiveness of our approach RLRP in both texture preservation and noise suppression tasks.

\begin{table}[h]
	\centering
	\caption{Restoration results for the case $\Phi = \bm{S}$(mask).}
	\resizebox{\linewidth}{!}{
		\begin{tabular}{c|ccccc}
			\hline
			\cline{2-6}
			Image & Method & SNR(dB) & Iter. & Time(s) & SSIM \\
			\hline
			\multirow{4}{*}{\textsf{Kodim\_wall}} 
			& CLRP & 13.519 & 224  & 14.22 & 0.537 \\
			& NYZ & 13.204 & 87 & 3.31 & 0.549\\
			& HKZ & 13.053 & 300 & 10.87 & 0.509\\
            & LWC & 15.132 & 24 & 9.58 & 0.621 \\
            & GAT & 11.509 & 193 & 23.12 & 0.556 \\
			& \textbf{RLRP} & \textbf{15.718} & 498  & 17.51 & \textbf{0.559} \\
			\hline
			\multirow{4}{*}{\textsf{Barbara\_RGB}} 
			& CLRP & 13.116 & 247  & 16.25 & 0.672 \\
			& NYZ & 12.853 & 92 & 3.49 & 0.696\\
			& HKZ & 12.605 & 300 & 10.89 & 0.649\\
            & LWC & 12.438 & 23 & 9.10 & 0.598 \\
            & GAT & 10.503 & 199 & 23.03 & 0.556 \\
			& \textbf{RLRP} & \textbf{15.984} & 490  & 18.60 & \textbf{0.756} \\
			\hline
	\end{tabular}}
	\label{result_mask}
\end{table}

Now, to evaluate the performance of our approach under random down-sampling, we select five representative images: (b), (e), and (h)–(j). First, we corrupt the original images with t-noise (intensity is $0.1$, degrees of freedom is $2$), followed by random down-sampling at a sampling probability of $0.4$. Unlike mask-based degradation, random down-sampling introduces more significant perturbations to the noise’s prior distribution, particularly as the down-sampling ratio increases.
For consistency, we employ identical model parameters across all test images: $\tau = 0.015$, $\mu = 0.2$, $c = 0.02$, $\beta = 2$, $\gamma = 1.6$, $r = 1$, $s = 2.01$, $\sigma=\eta = 0.4$, $\epsilon = 10^{-2}$.

\begin{figure*}[!t]
	\centering
	\includegraphics[width = 0.85\textwidth]{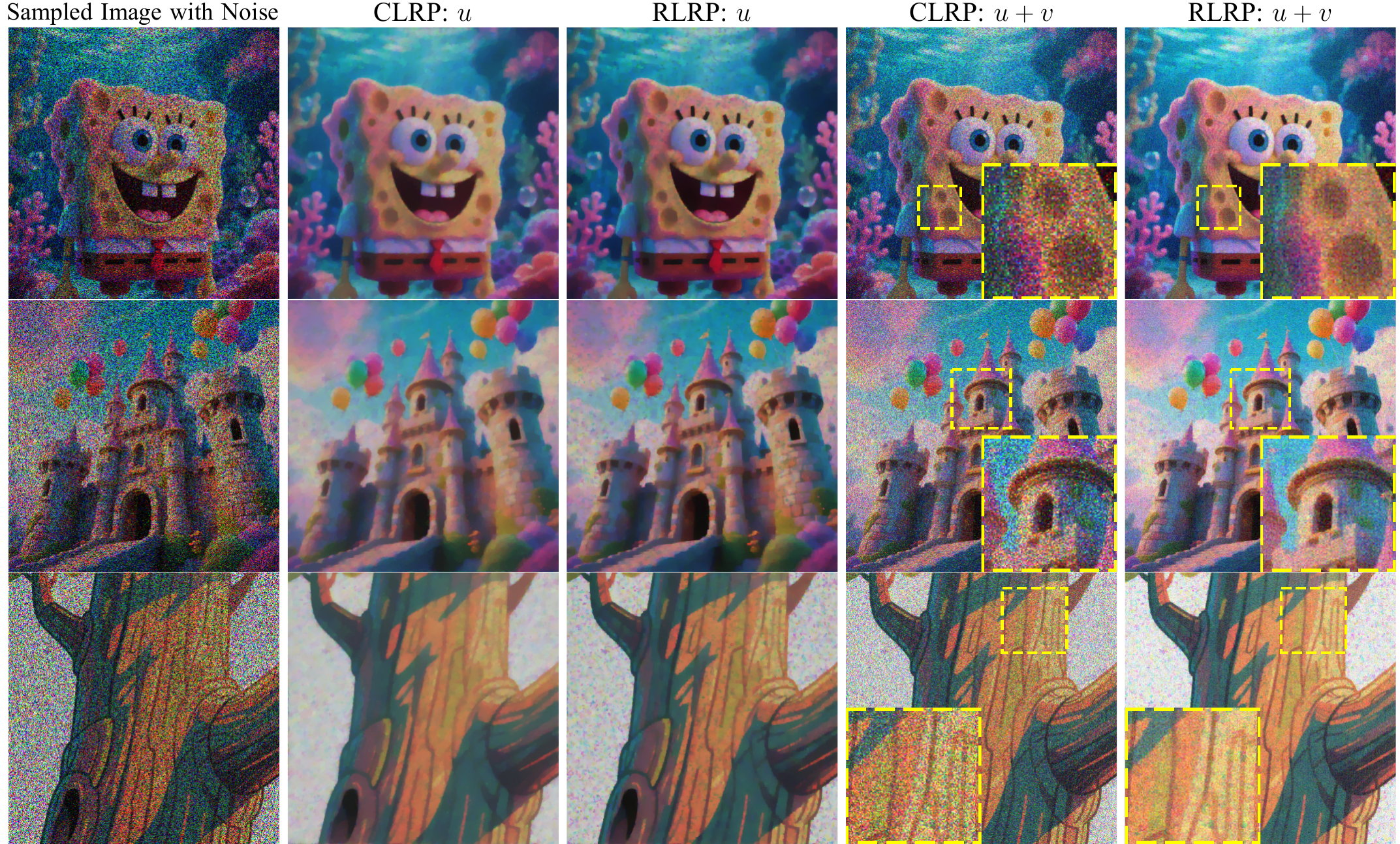}
	\caption{Results for downsampled cartoon images: $\Phi=\bm{S}$. From top to bottom: \textsf{Sponge\_Bob}, \textsf{Castle}, \textsf{Bole}. From left to right: the downsampled images, comparison of the component $u$, comparison of restored images. (left CLRP, right RLRP).}\label{cartoon-sample}
\end{figure*}

As illustrated in Figs. \ref{cartoon-sample}, under identical noise intensity and down-sampling conditions, our RLRP model achieves superior restoration of color images, yielding clearer textures with better structure and background separation. In contrast, the CLRP model suffers from noise contamination in the cartoon component, exhibiting noticeable outliers. 
These results demonstrate that our RLRP model effectively adapts to heavy-tailed noise distribution shifts induced by down-sampling, outperforming the CLRP approach.

Quantitative support is provided in Table \ref{result_sample}, where the  images reconstructed by our RLRP exhibit significantly higher SNR values than the other methods. Notably, for random down-sampling, the computational efficiency of RLRP remains competitive, with iteration counts and running time comparable to other cases, further underscoring its practical advantage.

\begin{table}[h]
	\centering
	\caption{Restoration results for the case $\Phi = \bm{S}$(downsample).}
	\resizebox{\linewidth}{!}{
		\begin{tabular}{c|ccccc}
			\hline
			\cline{2 - 6}
			Image & Method & SNR(dB) & Iter. & Time(s) & SSIM \\
			\hline
			\multirow{5}{*}{\textsf{Barbara}} 
			& CLRP & 13.316 & 41 & 4.22 & 0.451 \\
			& NYZ & 13.596 & 31 & 1.50 & 0.572\\
			& HKZ & 15.277 & 70 & 3.25 & 0.520\\
			& OMY & 14.654 & 35 & 14.34 & 0.505\\
            & LWC & 14.340 & 25 & 75.26 & 0.499 \\
            & GAT & 14.380 & 151 & 82.03 & 0.553 \\
			& \textbf{RLRP} & \textbf{17.150} & 57 & 3.69 & \textbf{0.593} \\
			\hline
			\multirow{4}{*}{\textsf{Kodim\_wall}} 
			& CLRP & 12.951 & 39 & 1.96 & 0.675 \\
			& NYZ & 11.186 & 30 & 0.98 & 0.622\\
			& HKZ & 13.764 & 69 & 2.06 & 0.699\\
            & LWC & 10.380 & 25 & 7.78 & 0.480 \\
            & GAT & 13.261 & 136 & 16.45 & 0.620 \\
			& \textbf{RLRP} & \textbf{16.790} & 61 & 2.16 & \textbf{0.756} \\
			\hline
			\multirow{4}{*}{\textsf{Sponge\_Bob}} 
			& CLRP & 8.339 & 30 & 9.68 & 0.286 \\
			& NYZ & 3.632 & 22 & 3.79 & 0.158\\
			& HKZ & 4.294 & 69 & 11.31 & 0.169\\
            & LWC & 7.475 & 30 & 91.84 & 0.244 \\
            & GAT & 9.759 & 153 & 85.78 & 0.351 \\
			& \textbf{RLRP} & \textbf{14.117} & 66 & 12.61 & \textbf{0.495} \\
			\hline
			\multirow{4}{*}{\textsf{Castle}} 
			& CLRP & 8.552 & 30 & 8.47 & 0.239 \\
			& NYZ & 3.719 & 22 & 4.09 & 0.121\\
			& HKZ & 4.380 & 68 & 10.20 & 0.137\\
            & LWC & 7.817 & 31 & 95.76 & 0.200 \\
            & GAT & 10.235 & 154 & 91.84 & 0.314 \\
			& \textbf{RLRP} & \textbf{14.753} & 67 & 11.73 & \textbf{0.435} \\
			\hline
			\multirow{4}{*}{\textsf{Bole}} 
			& CLRP & 8.648 & 31 & 9.89 & 0.242 \\
			& NYZ & 3.773 & 23 & 3.94 & 0.123\\
			& HKZ & 4.400 & 68 & 10.12 & 0.140\\
            & LWC & 7.865 & 33 & 101.60 & 0.214 \\
            & GAT & 10.324 & 155 & 90.01 & 0.312 \\
			& \textbf{RLRP} & \textbf{14.794} & 66 & 12.17 & \textbf{0.437} \\
			\hline
		\end{tabular}
	}
	\label{result_sample}
\end{table}

\begin{figure*}[!h]
	\centering
	\includegraphics[width = 0.85\textwidth]{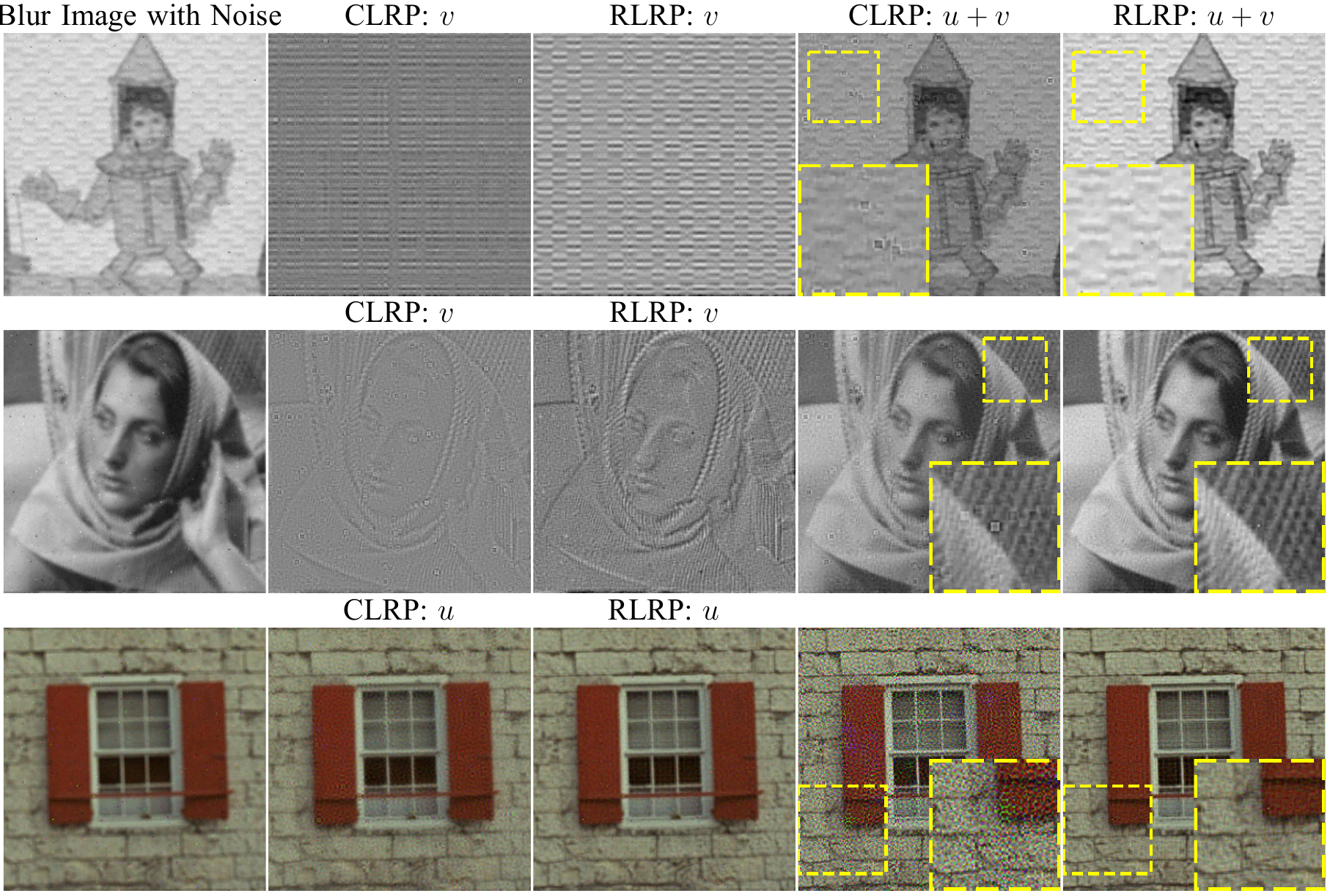}
	\caption{Results for blured images: $\Phi=\bm{B}$. From top to bottom: \textsf{Boy}, \textsf{Kodim\_wall}. From left to right: the downsampled images, comparison of the components (top $v$, bottom $u$), comparison of restored images. (left CLRP, right RLRP).}\label{normal-blur}
\end{figure*}

\begin{figure*}[h]
	\centering
	\includegraphics[width = 0.85\textwidth]{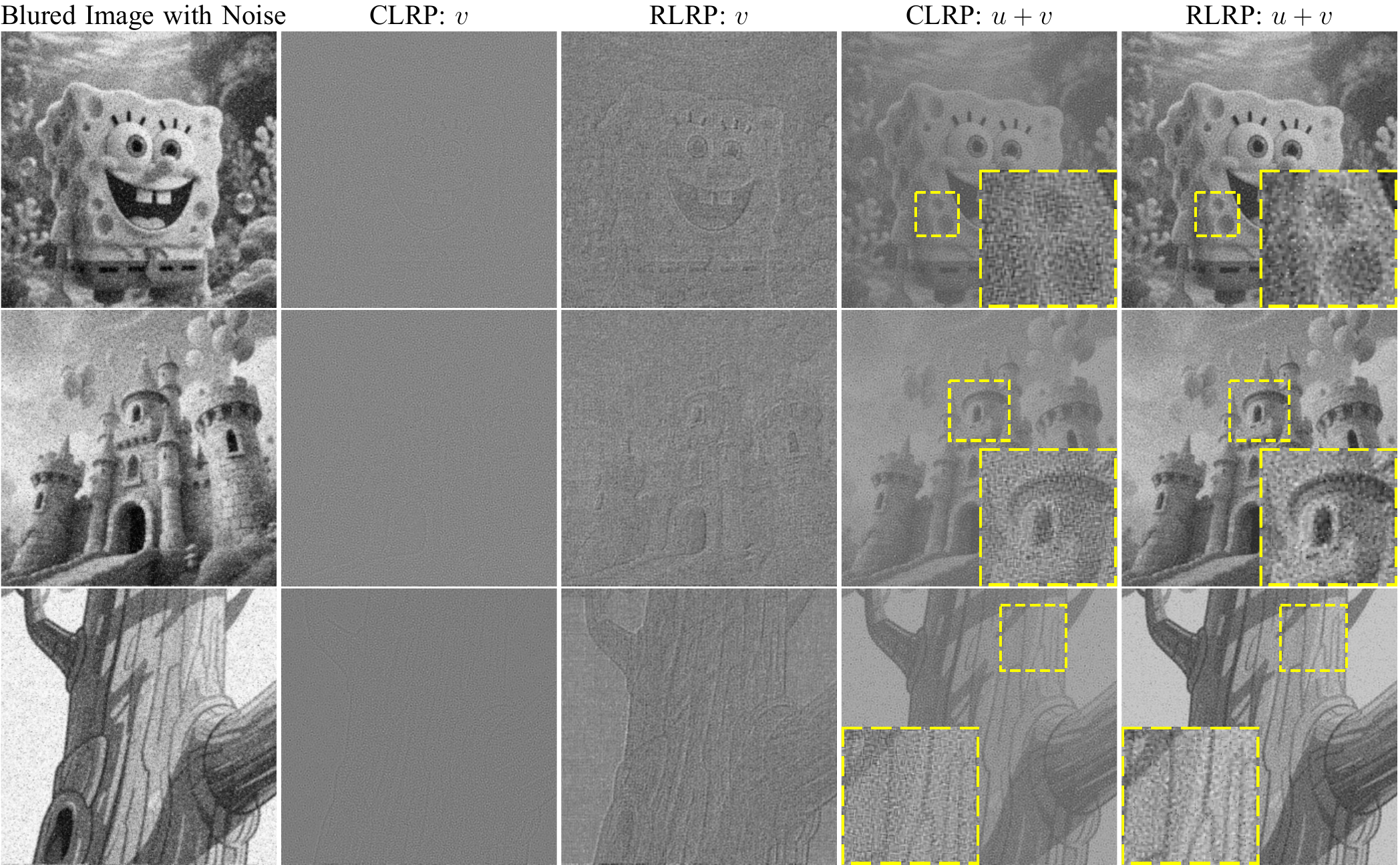}
	\caption{Results for blured gray cartoon images: $\Phi=\bm{B}$. From top to bottom: \textsf{Sponge\_Bob}, \textsf{Castle}, \textsf{Bole}. From left to right: the downsampled images, comparison of the component $v$, comparison of restored images. (left CLRP, right RLRP).}\label{cartoon-blur}
\end{figure*}

\subsection{Image deblurring}
In this subsection, we further consider the noise resistance experiment in which the known blurring operator $\Phi = B$ acts on the images. We evaluate our model using six test images: (a), (b), (e), and (h)-(j). Each image undergoes blurring through a $4\times 4$ average kernel. To comprehensively assess model performance under varying noise conditions, we apply t-distributed noise with different parameters: images (a), (b), and (e) degraded by noise with intensity 0.01 and 2 degrees of freedom, while images (h)-(j) are corrupted with stronger noise (intensity is 0.1). The efficacy of blurred-image restoration is particularly sensitive to noise levels \cite{BF12}. This is primarily because the blurring process modifies the intrinsic noise characteristics, such as by introducing a smoothing effect.
This phenomenon creates substantial challenges for the extraction of texture components from blurred images contaminated by heavy-tailed noise.

Regarding the model parameters, for the synthetic image (a) and cartoon-like images (h)-(j), we set $\tau = 10^{-3}$, $\mu = 0.1$, and natural images (b),(e), we set $\tau = 10^{-5}$, $\mu = 10^{-4}$. For our approach, we set $c = 0.03$, $\sigma=\eta = 0.6$. Regarding the algorithm of solving \eqref{CLRP}, we set $\gamma=1.6$, $r=1$, $s=2.01$. Our computational experiences show that penalty parameter $\beta$ is important for algorithmic implementation. 
For different images, (a) and (e) have large-scale structures, (b) only has local textures, and cartoon-type images (h)-(j) have only piece-wise smooth structures. The influence of noise on the restoration of textures in these images increases gradually. Therefore, setting the penalty parameter $\beta$ to $0.1$, $1.5$, and $2$ respectively is a reasonable way to balance the restoration difficulty, so that different algorithms can be compared on the premise of restoring high-quality images.

The restoration results are comprehensively presented in Figures \ref{normal-blur} and \ref{cartoon-blur}. Figure \ref{normal-blur} shows that when processing the synthetic image (a) under the combined effects of the blurring operator $B$ and heavy-tailed noise, the CLRP model produces significantly distorted texture components compared to the original image. This degradation primarily stems from two factors: (i) the model's strict adherence to low-rank constraints forces improper texture classification; (ii) noise interference disrupts the critical distinction between authentic low-rank textures and outlier components, ultimately compromising the accurate separation of cartoon and texture elements.
In contrast, our RLRP model exhibits markedly superior performance. Its restored results show three key advantages: (i) complete absence of visible noise artifacts in texture regions; (ii) effective resistance to the smearing effects typically induced by average blurring operations, and (iii) precise component separation that prevents cartoon patterns from appearing in texture regions. These improvements collectively demonstrate the model's enhanced capability to handle the challenging situations of simultaneous blurring and heavy-tailed noise contamination.
This phenomenon is more pronounced in the comparison of cartoon-like images presented in Fig. \ref{cartoon-blur}. For the CLRP model, the recovered texture is exceedingly faint, nearly imperceptible. 
This is caused by the repeated appearance of high-intensity noise that has not been eliminated in the image after multiple blurring processes.
In images with poorly defined textures, this effect may result in the signal intensity being lower than the noise intensity, corresponding to an SNR value less than zero ($\text{SNR} < 0$).

\begin{table}[h]
	\centering
	\caption{Results for image deblurring:$\Phi=\bm{B}$.}
	\resizebox{\linewidth}{!}{
		\begin{tabular}{c|ccccc}
			\hline
			\cline{2 - 6}
			Image & Method & SNR(dB) & Iter. & Time(s) & SSIM \\
			\hline
			\multirow{4}{*}{\textsf{Kodim\_wall}} 
			& CLRP & 5.371 & 53 & 2.67 & 0.457 \\
			& NYZ & 14.011 & 76 & 0.74 & 0.634\\
			& HKZ & 15.541 & 77 & 2.99 & 0.684\\
            & LWC & 16.960 & 6 & 1.05 & 0.731 \\
            & GAT & 13.549 & 81 & 10.41 & 0.459 \\
			& \textbf{RLRP} & \textbf{16.286} & 53 & 3.41 & \textbf{0.762} \\
			\hline
            \multirow{5}{*}{\textsf{Boy}} 
			& CLRP & 22.806 & 15 & 0.29 & 0.761 \\
			& NYZ & 21.479 & 76 & 0.74 & 0.653\\
			& HKZ & 24.781 & 50 & 0.44 & 0.781\\
			& OMY & 10.620 & 81 & 12.20 & 0.147 \\
            & LWC & 25.017 & 6 & 1.27 & 0.804 \\
            & GAT & 22.659 & 76 & 10.96 & 0.602 \\
			& \textbf{RLRP} & \textbf{27.453} & 41 & 0.89 & \textbf{0.870} \\
			\hline
			\multirow{5}{*}{\textsf{Barbara}} 
			& CLRP & 16.655 & 57 & 0.87 & 0.651 \\
			& NYZ & 16.020 & 77 & 0.74 & 0.597\\
			& HKZ & 17.844 & 54 & 0.53 & 0.706\\
			& OMY & 6.086 & 82 & 12.74 & 0.132 \\
            & LWC & 18.184 & 6 & 1.38 & 0.721 \\
            & GAT & 16.720 & 69 & 9.04 & 0.634 \\
			& \textbf{RLRP} & \textbf{18.600} & 51 & 1.25 & \textbf{0.722} \\
			\hline
			\multirow{5}{*}{\textsf{Sponge\_Bob}} 
			& CLRP & 5.364 & 57 & 0.87 & 0.199 \\
			& NYZ & 3.542 & 86 & 4.37 & 0.158\\
			& HKZ & 8.819 & 81 & 3.49 & 0.254\\
			& OMY & 2.241 & 42 & 35.27 & 0.067 \\
            & LWC & 8.167 & 6 & 6.94 & 0.244 \\
            & GAT & 13.037 & 67 & 55.99 & 0.405 \\
			& \textbf{RLRP} & \textbf{11.641} & 51 & 6.24 & \textbf{0.339} \\
			\hline
			\multirow{5}{*}{\textsf{Castle}} 
			& CLRP & 6.424 & 61 & 6.79 & 0.187 \\
			& NYZ & 4.575 & 86 & 4.37 & 0.138\\
			& HKZ & 9.844 & 81 & 3.61 & 0.227\\
			& OMY & 3.335 & 42 & 34.85 & 0.055 \\
            & LWC & 9.537 & 6 & 6.72 & 0.220 \\
            & GAT & 13.962 & 75 & 57.13 & 0.371 \\
			& \textbf{RLRP} & \textbf{12.661} & 51 & 6.60 & \textbf{0.306} \\
			\hline
			\multirow{5}{*}{\textsf{Bole}} 
			& CLRP & 7.833 & 61 & 5.90 & 0.210 \\
			& NYZ & 5.947 & 86 & 4.39 & 0.168\\
			& HKZ & 11.163 & 80 & 3.44 & 0.270\\
			& OMY & 5.035 & 42 & 34.16 & 0.073 \\
            & LWC & 10.983 & 6 & 6.79 & 0.248 \\
            & GAT & 14.616 & 78 & 57.70 & 0.398 \\
			& \textbf{RLRP} & \textbf{14.236} & 50 & 5.43 & \textbf{0.362} \\
			\hline
		\end{tabular}
	}
	\label{result_B}
\end{table}

All numerical results are summarized in Table \ref{result_B}. It is evident that, in the case of $\Phi = B$, the performance of our RLRP model exhibits the most significant contrast compared to other models. Particularly for low-rank and cartoon-like images, its SNR value is 6 or even up to 10 units higher than those of competing models. The presence of the blurring operator further illustrates that our RLRP model has effectively mitigated the impact of heavy-tailed noise and accurately recovered the underlying textures. Furthermore, it can be observed that the RLRP model solved via Algorithm \ref{alg_II} demonstrates comparable performance in terms of iteration count and computing time with other algorithms, and in some cases, it is even faster. 

\section{Conclusion}\label{sec_con}
We have developed a robust cartoon-texture decomposition model (RLRP in short) capable of handling heavy-tailed noise contamination. The proposed approach utilizes the Huber function for noise characterization and employs both splitting algorithms to efficiently solve the model under various linear degradation operators. Extensive numerical experiments demonstrate the proposed RLRP's superior noise resistance capabilities across different degradation scenarios, particularly outperforming existing methods in processing images with piecewise continuous smoothness and distinct low-rank texture patterns. While the current implementation shows promising results, one key limitation has been identified, i.e., comparative studies reveal that certain non-convex loss functions (e.g., Tukey's biweight) may offer superior performance to the Huber function. In the future, we will focus on the development of non-convex image decomposition models.

 \section*{Acknowledgment}
 This work was supported in part by National Natural Science Foundation of China (No. 12371303) and Zhejiang Provincial Natural Science Foundation of China at Grant No. LZ24A010001.

\ifCLASSOPTIONcaptionsoff
\newpage
\fi


\begin{thebibliography}{10}
	\providecommand{\url}[1]{#1}
	\csname url@samestyle\endcsname
	\providecommand{\newblock}{\relax}
	\providecommand{\bibinfo}[2]{#2}
	\providecommand{\BIBentrySTDinterwordspacing}{\spaceskip=0pt\relax}
	\providecommand{\BIBentryALTinterwordstretchfactor}{4}
	\providecommand{\BIBentryALTinterwordspacing}{\spaceskip=\fontdimen2\font plus
		\BIBentryALTinterwordstretchfactor\fontdimen3\font minus
		\fontdimen4\font\relax}
	\providecommand{\BIBforeignlanguage}[2]{{%
			\expandafter\ifx\csname l@#1\endcsname\relax
			\typeout{** WARNING: IEEEtran.bst: No hyphenation pattern has been}%
			\typeout{** loaded for the language `#1'. Using the pattern for}%
			\typeout{** the default language instead.}%
			\else
			\language=\csname l@#1\endcsname
			\fi
			#2}}
	\providecommand{\BIBdecl}{\relax}
	\BIBdecl
	
	\bibitem{BVSO03}
	M.~{Bertalmio}, L.~{Vese}, G.~{Sapiro}, and S.~{Osher}, ``Simultaneous
	structure and texture image inpainting,'' \emph{IEEE Trans. Image Process.},
	vol.~12, no.~8, pp. 882--889, 2003.
	
	\bibitem{FSBM10}
	M.~J. {Fadili}, J.~{Starck}, J.~{Bobin}, and Y.~{Moudden}, ``Image
	decomposition and separation using sparse representations: An overview,''
	\emph{Proc. IEEE}, vol.~98, no.~6, pp. 983--994, 2010.
	
	\bibitem{Meyer01}
	Y.~Meyer, \emph{Oscillating Patterns in Image Processing and Nonlinear
		Evolution Equations: The Fifteenth Dean Jacqueline B. Lewis Memorial
		Lecturs}.\hskip 1em plus 0.5em minus 0.4em\relax Boston: American
	Mathematical Society, 2001, vol.~22.
	
	\bibitem{ROF92}
	L.~Rudin, S.~Osher, and E.~Fatemi, ``Nonlinear total variation based noise
	removal algorithms,'' \emph{Physica D}, vol.~60, pp. 227--238, 1992.
	
	\bibitem{VO03}
	L.~A. Vese and S.~J. Osher, ``Modeling textures with total variation
	minimization and oscillating patterns in image processing,'' \emph{J. Sci.
		Comput.}, vol.~19, no. 1-3, pp. 553--572, 2003.
	
	\bibitem{NYZ13}
	M.~K. {Ng}, X.~{Yuan}, and W.~{Zhang}, ``Coupled variational image
	decomposition and restoration model for blurred cartoon-plus-texture images
	with missing pixels,'' \emph{IEEE Trans. Image Process.}, vol.~22, no.~6, pp.
	2233--2246, 2013.
	
	\bibitem{SO13}
	H.~Schaeffer and S.~Osher, ``A low patch-rank interpretation of texture,''
	\emph{SIAM J. Imaging Sci.}, vol.~6, no.~1, pp. 226--262, 2013.
	
	\bibitem{OMY14}
	S.~Ono, T.~Miyata, and I.~Yamada, ``Cartoon-texture image decomposition using
	blockwise low-rank texture characterization,'' \emph{IEEE Trans. Image
		Process.}, vol.~23, no.~3, pp. 1128--1142, 2014.
	
	\bibitem{ZH21}
	Z.~Zhang and H.~He, ``A customized low-rank prior model for structured
	cartoon-texture image decomposition,'' \emph{Signal Process. Image Commun.},
	vol.~96, p. 116308, 2021.

        \bibitem{PWH24}
        H.~Pan, Y.~Wen, and Y.~Huang, ``$L_0$ gradient-regularization and scale space representation model for cartoon and texture decomposition,'' \emph{IEEE Trans. Image
		Process.}, vol.~33, pp. 4016–4028, 2024.
	
	\bibitem{LWC25}
	K.~Li, Y.~Wen, and R.~H. Chan, ``Cartoon-texture image decomposition using
	least squares and low-rank regularization,'' \emph{J. Math. Imaging Vis.},
	vol.~67, no.~1, p. 5, 2025.

        \bibitem{GAT25}
	A.~Guennec, J.-F.~Aujol, and Y.~Traonmilin, ``Joint structure-texture low-dimensional modeling for image decomposition with a plug-and-play framework,'' \emph{SIAM J. Imaging Sci.}, vol.~18, no.~2, pp. 1344–1371, 2025.


	
	\bibitem{MMCAB19}
	M.~Mafi, H.~Martin, M.~Cabrerizo, J.~Andrian, A.~Barreto, and Adjouadi, ``A
	comprehensive survey on impulse and {G}aussian denoising filters for digital
	images,'' \emph{Signal Processing}, vol. 157, pp. 236--260, 2019.
	
	\bibitem{JC15}
	R.~Jennifer and M.~Chithra, ``Bayesian denoising of ultrasound images using
	heavy-tailed {L}evy distribution,'' \emph{IET Image Process.}, vol.~9, no.~4,
	pp. 338--345, 2015.
	
	\bibitem{Yan12}
	M.~Yan, ``Image and signal processing with non-{G}aussian noise: {EM}-type
	algorithms and adaptive outlier pursuit,'' Ph.D. dissertation, University of
	California, Los Angeles, 2012.
	
	\bibitem{Huber73}
	P.~Huber, ``Robust regression: Asymptotics, conjectures and {M}onte {C}arlo,''
	\emph{Ann. Stat.}, vol.~1, pp. 799--821, 1973.
	
	\bibitem{SZF20}
	Q.~Sun, W.~Zhou, and J.~Fan, ``Adaptive {H}uber regression,'' \emph{J. Am.
		Stat. Assoc.}, vol. 115, pp. 254--265, 2020.
	
	\bibitem{HHY16}
	L.~Hou, H.~He, and J.~Yang, ``A partially parallel splitting method for
	multiple-block separable convex programming with applications to robust
	pca,'' \emph{Comput. Optim. Appl.}, vol.~63, no.~1, pp. 273--303, 2016.
	
	\bibitem{CP11}
	A.~Chambolle and T.~Pock, ``A first-order primal-dual algorithm for convex
	problems with applications to imaging,'' \emph{J. Math. Imaging Vis.},
	vol.~40, pp. 120--145, 2011.
	
	\bibitem{cai2010singular}
	J.-F. Cai, E.~J. Cand{\`e}s, and Z.~Shen, ``A singular value thresholding
	algorithm for matrix completion,'' \emph{SIAM J. Optim.}, vol.~20, no.~4, pp.
	1956--1982, 2010.
	
	\bibitem{BV03}
	S.~Boyd and L.~Vandenberghe, \emph{Convex Optimization}.\hskip 1em plus 0.5em
	minus 0.4em\relax Cambridge University Press, 2003.
	
	\bibitem{HKZ15}
	D.~Han, W.~Kong, and W.~Zhang, ``A partial splitting augmented lagrangian
	method for low patch-rank image decomposition,'' \emph{J. Math. Imaging
		Vis.}, vol.~51, no.~1, pp. 145--160, Jan. 2015.
	
	\bibitem{t-distribution}
	H.~M. Walker and J.~Lev, ``Statistical inference,'' \emph{J.R. Stat. Soc. A. Stat.}, vol.~117, no.~266, p.~102, 1954.

	\bibitem{ssim}
	{Zhou Wang}, A.~Bovik, H.~Sheikh, and E.~Simoncelli, ``Image quality
	assessment: From error visibility to structural similarity,'' \emph{IEEE Trans. Image Process.}, vol.~13, no.~4, pp. 600--612, 2004.
	
	\bibitem{Arnold79}
	B.~C. Arnold, ``Some characterizations of the {C}auchy distribution,''
	\emph{Aust. J. Stat.}, vol.~21, no.~2, pp. 166--169, 1979.
	
	\bibitem{BT92}
	G.~E. Box and G.~C. Tiao, \emph{Bayesian Inference in Statistical
		Analysis}.\hskip 1em plus 0.5em minus 0.4em\relax New York: John Wiley and
	Sons, 1992.
	
	\bibitem{BF12}
	G.~Boracchi and A.~Foi, ``Modeling the performance of image restoration from
	motion blur,'' \emph{IEEE Trans. Image Process.}, vol.~21, no.~8, pp.
	3502--3517, 2012.
	
\end{thebibliography}
\end{document}